\documentclass{article}
\pdfoutput=1

     \usepackage[preprint, nonatbib]{neurips_2020}

\usepackage[utf8]{inputenc} 
\usepackage[T1]{fontenc}    
\usepackage{hyperref}       
\usepackage{url}            
\usepackage{booktabs}       
\usepackage{amsfonts}       
\usepackage{nicefrac}       
\usepackage{microtype}      

\usepackage{multirow}
\usepackage{graphicx, subcaption}
\usepackage{amssymb}
\usepackage{color}
\usepackage[dvipsnames]{xcolor}
\usepackage{makecell}

\usepackage{amsmath,amsfonts,bm}

\def\eqref#1{equation~\ref{#1}}

\def\1{\bm{1}}

\def\rvf{{\mathbf{f}}}

\def\rvv{{\mathbf{v}}}

\def\rvy{{\mathbf{y}}}

\def\ervd{{\textnormal{d}}}

\def\rmF{{\mathbf{F}}}

\def\rmM{{\mathbf{M}}}

\def\rmX{{\mathbf{X}}}
\def\rmY{{\mathbf{Y}}}

\def\ermI{{\textnormal{I}}}

\def\vy{{\bm{y}}}

\DeclareMathAlphabet{\mathsfit}{\encodingdefault}{\sfdefault}{m}{sl}
\SetMathAlphabet{\mathsfit}{bold}{\encodingdefault}{\sfdefault}{bx}{n}

\def\gC{{\mathcal{C}}}

\def\gF{{\mathcal{F}}}

\def\gH{{\mathcal{H}}}

\def\gL{{\mathcal{L}}}

\def\gX{{\mathcal{X}}}
\def\gY{{\mathcal{Y}}}

\def\sP{{\mathbb{P}}}

\def\sR{{\mathbb{R}}}

\DeclareMathOperator*{\argmax}{arg\,max}

\title{Alternating ConvLSTM: Learning Force Propagation with Alternate State Updates}

\author{%
  Congyue Deng \\
  Department of Mathematical Science \\
  Tsinghua University \\
  \texttt{dengcy16@mails.tsinghua.edu.cn}
  \And
  Tai-Jiang Mu \qquad\qquad\qquad\quad Shi-Min Hu \\
  Department of Computer Science and Technology
  \\
  Tsinghua University \\
  \texttt{\{taijiang, shimin\}@tsinghua.edu.cn}
}

\begin{document}

\maketitle

\begin{abstract}

Data-driven simulation is an important step-forward in computational physics when traditional numerical methods meet their limits.
Learning-based simulators have been widely studied in past years; however, most previous works view simulation as a general spatial-temporal prediction problem and take little physical guidance in designing their neural network architectures.
In this paper, we introduce the \emph{alternating} convolutional Long Short-Term Memory (Alt-ConvLSTM) that models the force propagation mechanisms in a deformable object with near-uniform material properties.
Specifically, we propose an accumulation state, and let the network update its cell state and the accumulation state alternately.
We demonstrate how this novel scheme imitates the alternate updates of the first and second-order terms in the forward Euler method of numerical PDE solvers.
Benefiting from this, our network only requires a small number of parameters, independent of the number of the simulated particles, 
and also retains the essential features in ConvLSTM, making it naturally applicable to sequential data with spatial inputs and outputs.
We validate our Alt-ConvLSTM on human soft tissue simulation with thousands of particles and consistent body pose changes.
Experimental results show that Alt-ConvLSTM efficiently models the material kinetic features and greatly outperforms vanilla ConvLSTM with only the single state update.

\end{abstract}

\section{Introduction}

Physical simulation plays an important role in various fields such as computer animation \cite{fedkiw2001visual, foster2001practical, losasso2008two, stomakhin2013material}, mechanical engineering \cite{birdsall1991particle}, and robotics \cite{clegg2015animating, clegg2018learning, li2018learning, erickson2019assistive, clegg2020learning, xiang2020sapien}.
In the past decades, numerical partial differential equation (PDE) solvers have been the most prevalent technique for physical simulation tasks with their theoretical foundations and outstanding performances \cite{brackbill1988flip, birdsall1991particle, osher2004level, gibou2018review, liu2010smoothed, macklin2013position, stomakhin2013material}.
However, these numerical methods leave behind several unsolved problems, including the laborious computations and the difficulty to model complex materials.
Data-driven methods, especially learning-based methods, then become an important step-forward, as they provide instant feedbacks at run time and can summarize material properties from the underlying data \cite{kim2017data, pai2018human}.
However, new challenges emerge as the old ones leave.
One limitation that impedes learning-based methods from performing better simulations is the trade-off between model complexity and representability.
In traditional numerical methods, a model with certain number of parameters can be applied to physical systems of any scale as long as it is established for a material, but here in a neural network, the number of parameters is always growing as the simulation task goes to larger scales.
We attribute this to the neglect of physical guidance in neural network design.
In most previous works, only the spatial-temporal structure of networks are fine tuned, while the physical meaning in each computation step is not carefully considered.
%
%


%


In this paper, we show that this trade-off can be avoided by making a neural network more "physical".
We propose the \emph{alternating} convolutional Long Short-Term Memory (Alt-ConvLSTM) 
for simulating deformable objects with near-uniform material kinetic properties.
Inspired by the the alternate updates of the first and second-order terms in the forward Euler method of numerical PDE solvers, we
extend the classical ConvLSTM to fully convolutional operations with an additional accumulation state and let it update its cell state and the accumulation state alternately.
This novel adaptation efficiently models the force propagation process within complex physical systems and is heuristically adhered to certain kinetic laws.
%
We also generalize the Alt-ConvLSTM to non-Euclidean domains with graph convolutions and 
assemble Alt-ConvLSTM cells into an encoding-decoding architecture.
%
We evaluate our model on the Dynamic FAUST dataset \cite{bogo2017dynamic} for simulating human soft tissue dynamics with respect to body movements, and results show that our Alt-ConvLSTM is capable of faithfully and efficiently simulating the dynamics of complex physical systems compared to the vanilla ConvLSTM and existing methods.

In summary, we make the following contributions:
\begin{itemize}
    \item \emph{Physical interpretations.}
    To the best of our knowledge, we are of the first attempts to design a neural network structure under full physical guidance and thus make the network interpretable as well as capable of achieving more realistic simulation.
    %
    %
    %
    \item \emph{Strong representability with low complexity.} We adopt a fully convolution temporal structure, resulting in $O(1)$ network parameters \textit{w.r.t.} an $n$-particle system.
    Despite its low complexity, experiments show that Alt-ConvLSTM performs strong representability.
\end{itemize}
%
%
%
%

\section{Related Work}

\paragraph{Learning-based physical simulation}
Machine learning is widely used in data-driven physical simulation to infer trajectories, deformations, and interactions, but the trade-off between model complexity and representability long exists before it can be applied to large-scale physical systems.
Beside the simplest case of simulating one single rigid body \cite{RempeDynamics2020}, initial attempts in learning multi-object interactions \cite{chang2016compositional, battaglia2016interaction, kipf2018neural, van2018relational} can only simulate a few objects/particles since they are maintaining a fully connected interaction graph.
More recent works push this limit to tens \cite{li2019propagation} or hundreds \cite{mrowca2018flexible, li2018learning} of particles by introducing sparser graph representations.
For deformable objects, reducing the number of particles (vertices) directly is infeasible, but simplifications on the physical system are still carried out in the means of data encoding \cite{casas2018learning, santesteban2019learning, santesteban2020softsmpl} or timeline truncation \cite{jin2018pixel, geng2020coercing}, which in turn introduces additional errors and makes these methods heavily problem-dependent.
%
%
%
%
Besides, neural networks can also be plugged into traditional numerical schemes, putting corrections \cite{wu2020recovering}, enriching physical details \cite{lahner2018deepwrinkles}, forming a hybrid model \cite{kim2017data, schenck2018spnets}, or serving as an auxiliary tool \cite{lee2019efficient}.
However, hardly any existing models adopt an explicit physical guidance in its neural network structure like us.

\paragraph{Spatial-temporal data processing}
We also get inspirations from the broader topics of spatial-temporal data processing.
%
%
%
Attempts have been widely made on extending convolutional neural networks (CNNs) to perform sequence-to-sequence learning.
\cite{chan2019everybody,wang2018video} apply residual networks (ResNet) to video-to-video transfer,
but their output frames are synthesized individually with historical information neglected.
\cite{zhang2017deep} predicts crowd flows with ResNet by fusing short-term, long-term, and periodic predictions made separately, but this way of adding temporal dependencies heavily relies on crowd behavioral priors and can be hardly generalized to other tasks.
Recurrent CNN \cite{liang2015recurrent} adds a time-evolving hidden state to the convolution layer, making the network applicable to sequential data while processing spatial information.
Convolutional LSTM \cite{xingjian2015convolutional, zeng2017recurrent} further adopts the recurrence in LSTM.
Our method is closely related to these works.
In the meantime, we focus on the spatial structures of fully convolutional networks \cite{badrinarayanan2015segnet,long2015fully, ronneberger2015u} in favor of their small parameter amounts.

\section{Preliminaries}

\subsection{Problem Formulation}

Suppose we have a physical system over graph domain $G=\{V,E\}$; the graph nodes $V$ are particles, and the edges $E$ are the interactive relations between adjacent particles.
Denote the states of all particles at time $t$ as $\rmY_t\in\sR^{K_Y\times |V|}$, the applied external perturbations as $\rmF_t\in\sR^{K_F\times |V|}$, where $K_Y,K_F$ are feature dimensions for each particle.
%
%
For simulation, we hope to predict the physical state $\rmY_t$ at time $t$ given current external perturbation $\rmY_t$ and all past states $\rmY_1,\cdots,\rmY_{t-1}$ by maximizing the following probability
\begin{equation}
    \hat{\rmY}_t = \argmax_{\rmY_t}\sP(\rmY_t ~|~ \rmF_t, \rmY_1,\cdots,\rmY_{t-1}).
\end{equation}
%
%
We further make a near-uniform material assumption and formulate it in the following way:
For all nodes $v\in V$, their states $\rmY_t(v)$ can be approximated by a function $f$ with shared parameters $\theta$
\begin{equation}
    \hat{\rmY}_t(v) =f_\theta(\rmF_t,\rmY_1,\cdots,\rmY_{t-1}; N(v)),
    \label{eq:pd-2}
\end{equation}
where $N(v)$ is a subgraph of $G$ containing node $v$ and sampled according to a certain rule; for example, $N(v)$ can be the one-ring neighbourhood of $v$.
$f_\theta$ is inherently a spatial-temporal function of physical states, external perturbations and local particle interactions,
giving feasibility to fully convolutional recurrent models.


%
%
%
%

\subsection{The Forward Euler Method}
The forward Euler method is widely used to perform state updates in physical systems and can be adopted to solve the problem mentioned above.
For each particle, given the known positions $\rvy_{t-1}$ and velocities $\rvv_{t-1}$ at timestep $t-1$, a simulator aims to determine the new positions $\rvy_{t}$ and velocities $\rvv_{t}$ at timestep $t$.
Newton's second law indicates
\begin{equation}\label{eq:newton_second_law}
    \frac{\ervd}{\ervd t}
    \begin{pmatrix}
    \rvy \\
    \rvv
    \end{pmatrix}
    =
    \begin{pmatrix}
    \rvv \\
    \rmM^{-1} \left(\rvf^{ex} + \rvf^{in}(\rvy, \rvv) \right)
    \end{pmatrix},
\end{equation}
where $\rmM$ is the mass matrix, $\rvf^{ex}$ is the external force, and $\rvf^{in}$ is the internal force.
The explicit forward Euler method approximates this with step size $h$ by
\begin{equation}\label{eq:forward_euler}
    \begin{pmatrix}
    \Delta \rvy_{t} \\
    \Delta \rvv_{t}
    \end{pmatrix}
    = h
    \begin{pmatrix}
    \rvv_{t-1} \\
    \rmM^{-1} \left(\rvf_{t-1}^{ex} + \rvf^{in}(\rvy_{t-1}, \rvv_{t-1}) \right)
    \end{pmatrix},
\end{equation}
and alternately updates the velocities (second-order term) 
with $\rvv_t = \rvv_{t-1}+\Delta\rvv_t$ and the positions (first-order term) 
with $\rvy_t = \rvy_{t-1}+\Delta\rvy_t$.
In Sec.~\ref{sec:alt_convlstm_cell}, we will show how to incorporate the forward Euler updates into the network for predicting physical dynamics.

\section{The Model}
\label{sec:model}

We now present our alternating ConvLSTM with a structure design guided by the forward Euler method.
%
We first give the Alt-ConvLSTM formulation based on alternate state updates in the standard 2D case and demonstrate how it models force propagation (Sec.~\ref{sec:alt_convlstm_cell}).
%
Then, we generalize it to non-Euclidean domains with graph convolutions (Sec.~\ref{sec:graph_convolution}).
%
We finally assemble multiple Alt-ConvLSTM cells into an encoding-decoding network for spatial-temporal simulation (Sec.~\ref{sec:architecture}).

\subsection{Alternating ConvLSTM Cell}
\label{sec:alt_convlstm_cell}

%
We now introduce the update rules in a single Alt-ConvLSTM cell. For simplicity, we first do this on regular grids with the standard 2D convolution. 
As in most LSTM formulations, we denote by $\gX_1,\cdot,\gX_T$ the cell inputs, $\gC_1,\cdots,\gC_T$ the cell states, $\gH_1,\cdots,\gH_T$ the hidden states, and $i_t,f_t,o_t$ the input gate, forget gate, output gate respectively. These are all 3D tensors with their last two dimensions preserving the spatial structure.
We define an alternating ConvLSTM cell to be
\begin{equation}\label{eq:alt_convlstm}
\begin{split}
    i_t &= \sigma(W_{xi}*\gX_t + W_{hi}*\gH_{t-1} + W_{ci}*\gY_{t-1} + b_i), \\
    f_t &= \sigma(W_{xf}*\gX_t + W_{hf}*\gH_{t-1} + W_{cf}*\gY_{t-1} + b_f), \\
    \gC_t &= f_t\circ\gC_{t-1} + i_t\circ\tanh(W_{xc}*\gX_t + W_{hc}*\gH_{t-1} + b_c), \\
    \gY_t &= \gY_{t-1} + \gC_t, \\
    o_t &= \sigma(W_{xo}*\gX_t + W_{ho}*\gH_{t-1} + W_{co}*\gY_t + b_o), \\
    \gH_t &= o_t\circ\tanh(\gC_t). \\
\end{split}
\end{equation}
Compared with the vanilla ConvLSTM, there are two major modifications.
First, we add a cell accumulation state $\gY_t = \sum_{\tau=1}^t \gC_\tau$ and let the three gates look at this accumulation state instead of a single cell state $\gC_t$ at the peephole connections $W_{c\{i,f,o\}}$.
%
%
Letting the cell state $\gC_t$ be the velocities and the accumulation state $\gY_t$ be the positions, the update rule of these two states (line 3-4 in Eq. \ref{eq:alt_convlstm}) matches with the alternate second and first-order updates in the forward Euler method in Eq. \ref{eq:forward_euler},
with only a slight difference in the presence of $i_t,f_t$ that can be viewed as energy losses during force propagation.
Second, peephole connections themselves are modified from element-wise multiplications to convolutions,
as convolutions on particle positions encode the neighbourhood geometries, which is more relevant to force propagation than element-wise multiplications.

\begin{figure}[t]
    \centering
    \begin{subfigure}[t]{0.24\textwidth}
        \centering
        \includegraphics[width=\linewidth, trim=80 100 70 100, clip]{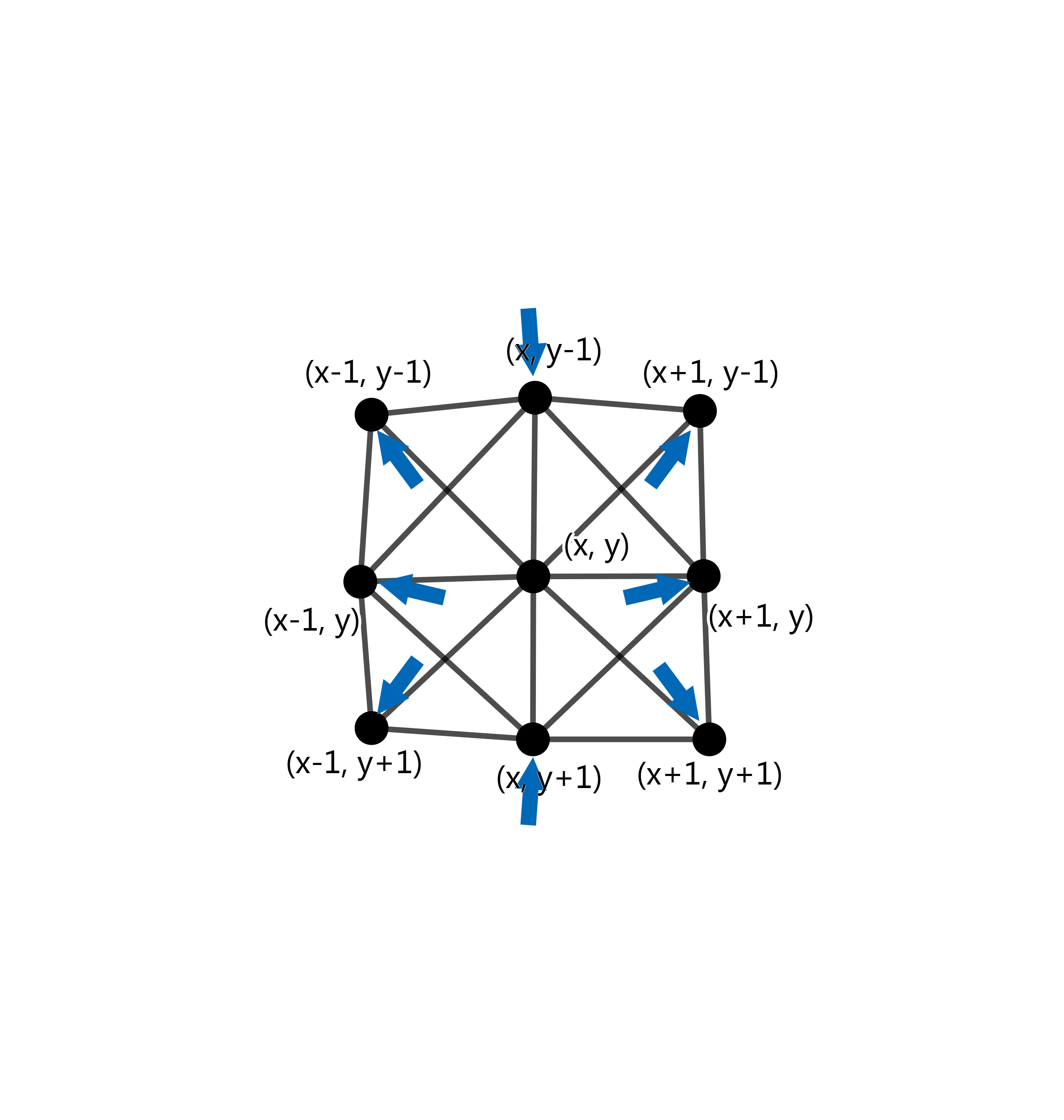}
        \caption{$t-1$ initial state}
        \label{fig:spring-ball_1}
    \end{subfigure}
    \begin{subfigure}[t]{0.24\textwidth}
        \centering
        \includegraphics[width=\linewidth, trim=80 100 70 100, clip]{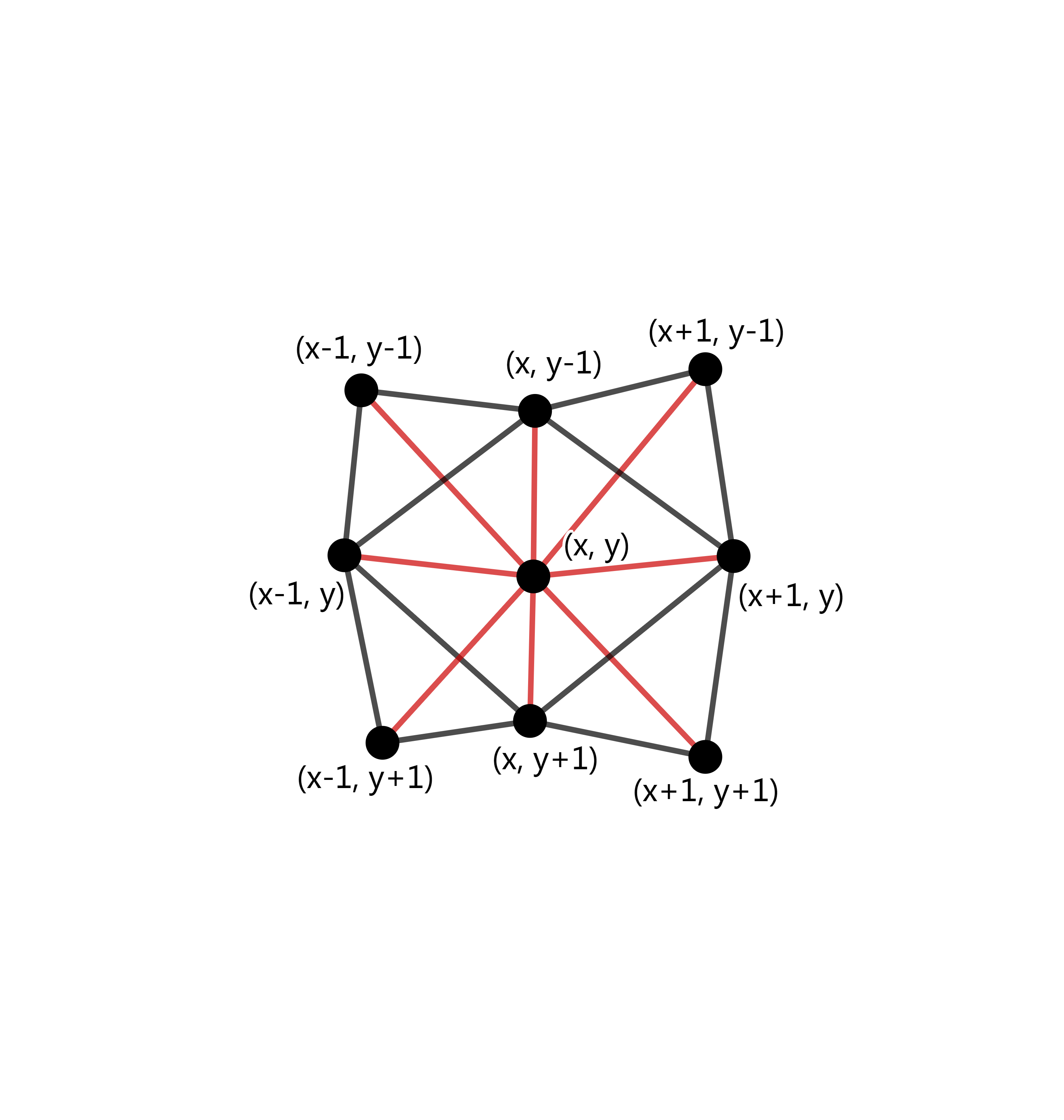}
        \caption{$t-1$ update}
        \label{fig:spring-ball_2}
    \end{subfigure}
    \begin{subfigure}[t]{0.24\textwidth}
        \centering
        \includegraphics[width=\linewidth, trim=80 100 70 100, clip]{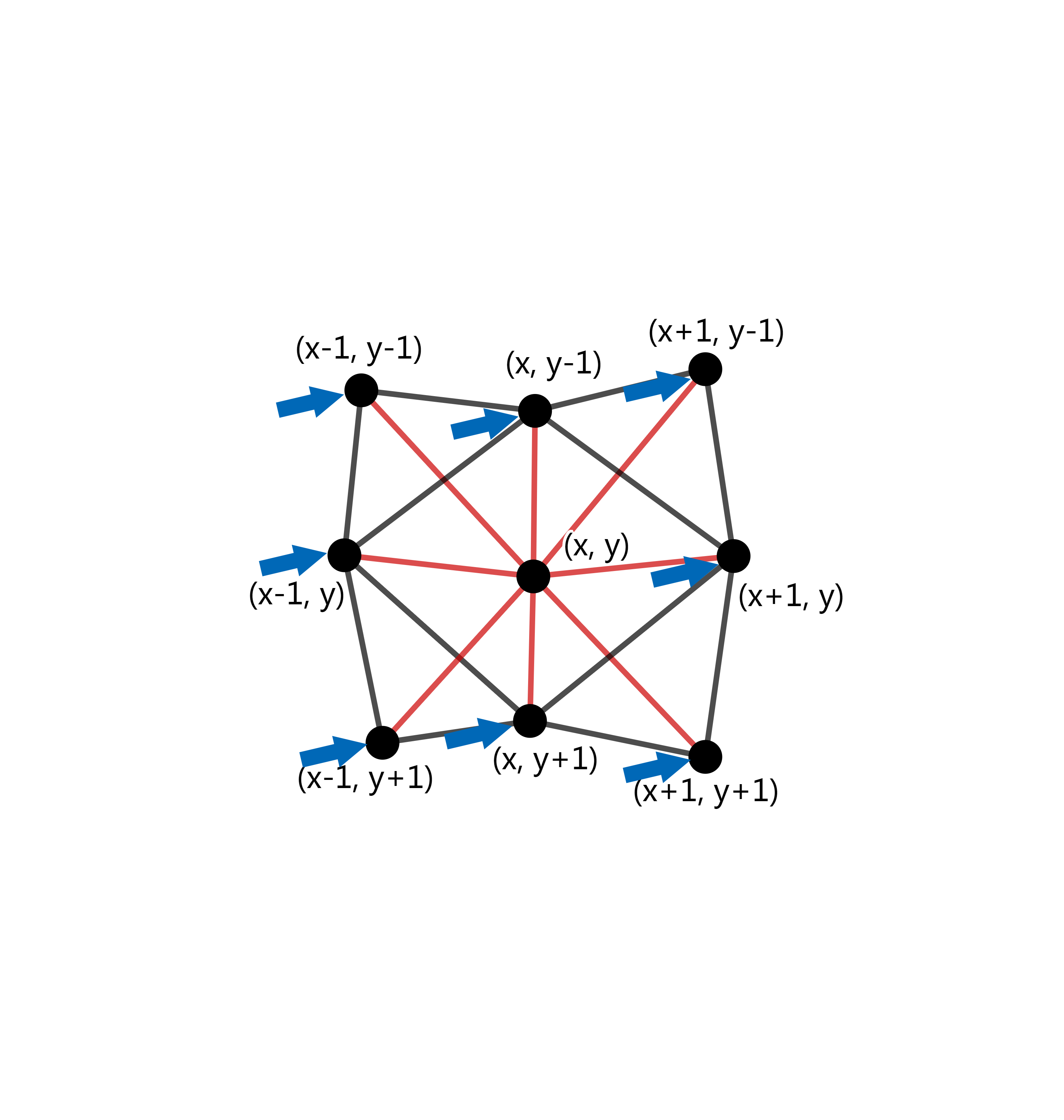}
        \caption{external force at $t$}
        \label{fig:spring-ball_3}
    \end{subfigure}
    \begin{subfigure}[t]{0.24\textwidth}
        \centering
        \includegraphics[width=\linewidth, trim=80 100 70 100, clip]{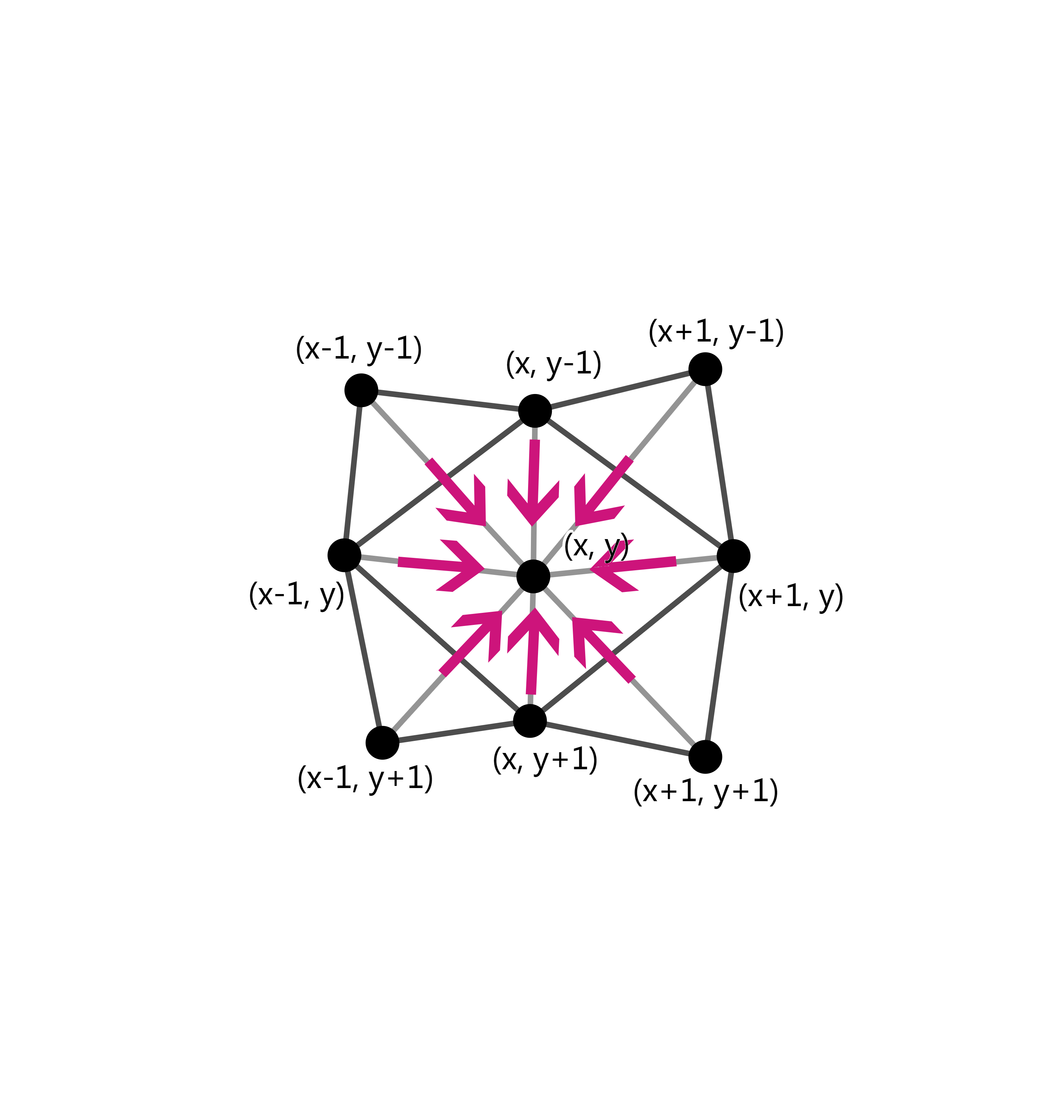}
        \caption{force propagation at $t$}
        \label{fig:spring-ball_4}
    \end{subfigure}
    \caption{Force propagation and state update around particle $(x,y)$ modeled by Alt-ConvLSTM.
    }
    \label{fig:spring-ball}
\end{figure}

More concretely, the Alt-ConvLSTM cell models the force propagation process inside the object in the following way:
At timestep $t$, cell input $\gX_t$ indicates that an external force is applied to the particles (Fig. \ref{fig:spring-ball_3} {\color{blue}blue}).
There is also an internal force caused by the position changes of neighbouring particles at time $t-1$ (Fig. \ref{fig:spring-ball_2} {\color{red}red}), which is encoded in $\gH_{t-1}$.
The total force $\gF_t(x,y)$ applied to the whole neighbourhood of particle $(x,y)$ is then represented by the tensor
\begin{equation}
    \gF_t(x,y) = \tanh(W_{xc}*\gX_t(x,y) + W_{hc}*\gH_{t-1}(x,y) + b_c).
\end{equation}
Assuming uniform mass distribution, the force is equivalent to the acceleration (up to a scalar), and thus we can update particle velocities with $\gC_t = \gC_{t-1} + \gF_t$.
For non-idealized materials, there are also energy losses when velocity $\rvv_{t-1}$ retains over time and when neighbourhood force $\gF_t(x,y)$ propagates to $(x,y)$ (Fig. \ref{fig:spring-ball_4} {\color{magenta}magenta}), due to the frictions or any other intricate mechanisms.
These dampening effects can always be modeled with discounting factors, which are the forget gate $f_t$ and the input gate $i_t$ here.
This gives us the full cell state update
\begin{equation}
    \gC_t = f_t\circ\gC_{t-1} + i_t\circ\gF_t,
\end{equation}
followed by the accumulation state update $\gY_t = \gY_{t-1} + \gC_t$.
Finally, the internal force caused by current position update is encoded in $\gH_t$ with factor $o_t$ and left for further use at time $t+1$.
This can be understood by an analogy to the simplest 1D spring case, where the potential energy $E$, particle positions $\vy,\vy_0$, and the elastic coefficient $k$ approximately follow the Hooke's law $E \sim k(\vy-\vy_0)^2$, giving the tension $F$ approximation
\begin{equation}\label{eq:hooke_law}
    F\Delta\vy = \Delta E = 2k(\vy-\vy_0)\Delta\vy + O(\Delta\vy^2). 
\end{equation}
Comparable to $(\vy-\vy_0)\Delta\vy$ in Eq.~\ref{eq:hooke_law}, the $o_t\circ\tanh(\gC_t)$ encoding is also a multiplication of the geometric structure $W_{co}*\gY_t$ and the velocity $\gC_t$ (up to some nonlinearities), and thus holds a good representability on force approximation.

An accompanying benefit of this Alt-ConvLSTM is the apparent reduction in model parameters.
For a physical system of $n$ particles, classical LSTM needs $O(n^2)$ parameters because of the matrix multiplications, vanilla ConvLSTM with Hadmard peephole connections needs $O(n)$ parameters, while our Alt-ConvLSTM only needs $O(1)$ parameters as it is fully convolutional.
This greatly saves GPU memory and makes learning large scale dynamics more tractable.

\subsection{Plug in Graph Convolutions}
\label{sec:graph_convolution}

Since the majority of deformable objects are represented on non-Euclidean domains such as interaction graphs or triangular meshes, we adapt our Alt-ConvLSTM for such input data.
A straightforward way is to replace the standard 2D convolutions in Alt-ConvLSTM (Eq. \ref{eq:alt_convlstm}) with graph convolutions.
Since graph convolution can also be viewed as message passing between nodes via edges, it perfectly fits into our intention of modeling force propagation.
Specifically, we adopt the graph convolution proposed in \cite{kipf2016semi}
\begin{equation}
    \textnormal{Conv}(\gX,\Theta) = \hat{D}^{-1/2}\hat{A}\hat{D}^{-1/2}\gX\Theta,
\end{equation}
where $\hat{A} = A+\ermI$ is the binary graph adjacency matrix with inserted self-loops, $\hat{D}_{ii} = \sum_{j=1}^{|V|} \hat{A}_{ij}$ is the diagonal degree matrix, and $\Theta$ is the learnable weight matrix.
%
%
We adopt this edge-weight-independent graph convolution because, in a dynamical system, edge weights are inconstant over time while the graph topology stays invariant.
As a compensation, we feed the particle positions into the network at each timestep, which enables the network to look at the graph geometry \cite{kostrikov2018surface}.

\subsection{Encoding-Decoding Architecture}
\label{sec:architecture}

\begin{figure}[t]
    \centering
    \includegraphics[width=0.95\linewidth]{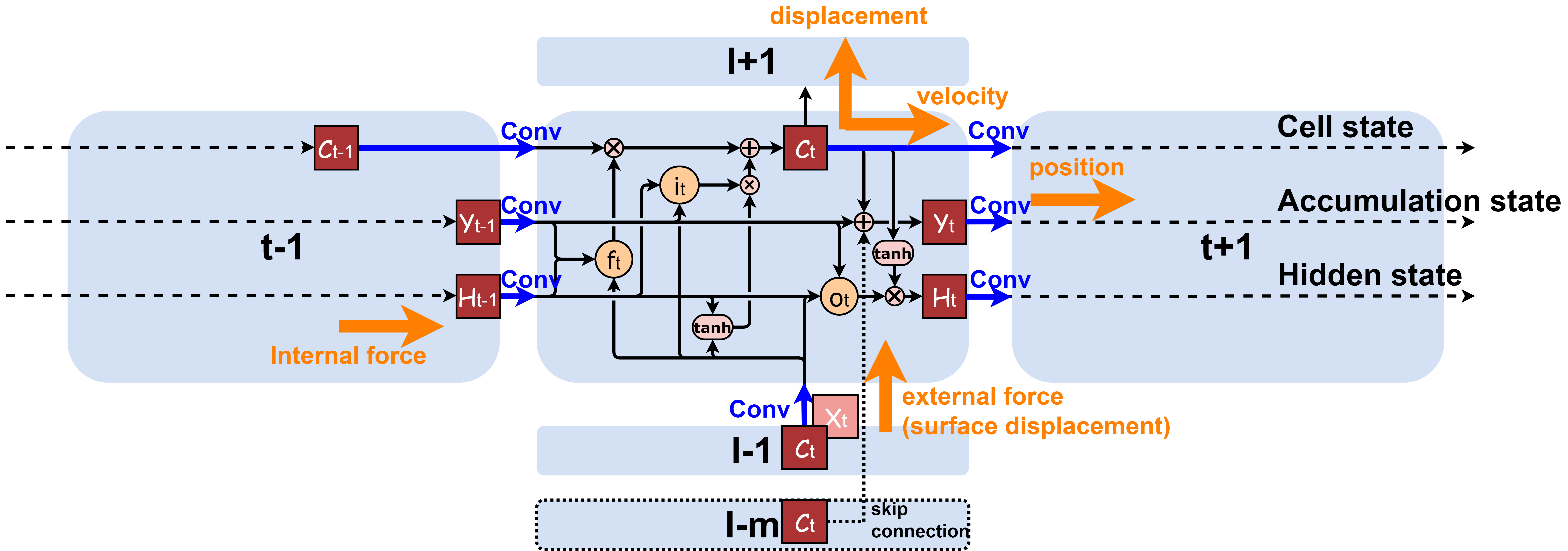}
    \caption{Alt-ConvLSTM structure {for physical simulation. }}
    \label{fig:network_architecture}
\end{figure}

Like any convolutional structures, Alt-ConvLSTM cells can also be stacked into multi-layers, 
assembling a more complex neural network architecture, where the input of layer $l$ is the cell state of layer $l-1$, \textit{i.e.} $\gX_t^{l} = \gC_t^{l-1}$.
This holds the meaning that the position change at layer $l-1$ causes an external force for layer $l$, just as in computational solid mechanics, perturbations are sometimes represented as displacements on surfaces.
We further let the number of channels for the layers be first increasing and then decreasing, which yields an encoding-decoding network architecture with the consecutive layers of increasing channels the encoder and those of decreasing channels the decoder.
%
%
%
%
%
%
From the perspective of the message passing, while one single Alt-ConvLSTM cell allows message passing from each node to their neighbouring nodes and models the force propagation within one-ring neighbourhoods, stacked layers model the long-distance propagation in larger neighbouring areas.

We further allow skip connections between nonadjacent layers $l$ and $l-m$ by
\begin{equation}
    \gY_t^l = \gY_{t-1}^l + \gC_t^l + \gC_t^{l-m}.
\end{equation}
This can be interpreted as velocity decomposition, where $\gC_t^{l-m}$ is the rigid motion in global coordinates and $\gC_t^l$ is the soft motion in local coordinates.
Moreover, it also enables the network to capture higher-frequency features, as consecutive convolutions smooths input features by taking averages.
The complete network structure is shown in Figure \ref{fig:network_architecture}, with the physical meanings marked in {\color{orange}orange}.

\section{Experiments}

In this section, we demonstrate the capability of simulating complex dynamics of our proposed Alt-ConvLSTM on a public available human motion dataset.

\subsection{Setup}

\textbf{Dataset.}
We test our model on the Dynamic FAUST Dataset \cite{bogo2017dynamic} consisting of 129 scanned 4D sequences of human motions with soft tissue movement captured at 60 fps.
Each frame contains a tissue surface mesh $\rmY_t$ of 6,890 vertices and a skeletal pose with body shape parameters which can be converted into an SMPL \cite{loper2015smpl} mesh model $\rmX_t$ of 6,890 vertices.
All tissue and SMPL meshes are in vertex correspondence, and thus can be represented on a constant graph domain $G$ with $|V| = $ 6,890 nodes.
%
Our goal is to predict the vertex coordinates of the tissue mesh $\rmY_t$ at every timestep $t$, with past and current body poses $\rmX_1,\cdots,\rmX_t$ and all past tissue states $\rmY_1,\cdots,\rmY_{t-1}$.
We treat human soft tissue as a near-uniform material
for three reasons: first, it has a near-constant mass density which is approximately the density of water; second, 
different parts of body tissue are closely linked and no relative sliding occurs between them; third, its response to forces is similar among all body parts.
%
We divide the 129 sequences into a training set of 80 sequences, a validation set of 20 sequences, and a test set of 29 sequences after shuffling the dataset.

\begin{table}[t]
  \caption{Future prediction error on known motions}
  \label{tab:future_prediction}
  \centering
  \begin{tabular}{llllll}
    \toprule
    \multirow{2}{*}{\makecell[c]{Error\\type}} & \multirow{2}{*}{Model} & \multicolumn{4}{c}{Per-vertex error: mean$\pm$sd (mm)} \\
    \cmidrule(r){3-6}
    && 20 steps & 30 steps & 40 steps & 50 steps \\
    \midrule
    & SMPL & 17.76$\pm$2.67 & 17.87$\pm$2.67 & 18.07$\pm$2.66 & 18.26$\pm$2.69 \\
    \midrule
    \multirow{6}{*}{\makecell[c]{Single\\-step}} & ConvLSTM-CP-$\Delta Y$ & 21.44$\pm$20.67 & 41.06$\pm$32.54 & 61.09$\pm$42.91 & 77.62$\pm$49.00 \\
    & ConvLSTM-NP-$\Delta Y$ & 21.21$\pm$20.52 & 40.53$\pm$32.29 & 60.36$\pm$42.70 & 76.76$\pm$48.89 \\
    & ConvLSTM-CP-$Y$ & 2.50$\pm$0.38 & 2.23$\pm$0.45 & 2.19$\pm$0.48 & 2.58$\pm$0.52 \\
    & ConvLSTM-NP-$Y$ & 3.21$\pm$0.43 & 2.98$\pm$0.46 & 2.95$\pm$0.50 & 2.95$\pm$0.51 \\
    & \textbf{Alt-ConvLSTM} & \textbf{0.97$\pm$0.41} & \textbf{1.09$\pm$0.48} & \textbf{1.21$\pm$0.53} & \textbf{1.30$\pm$0.56} \\
    \midrule
    \multirow{6}{*}{\makecell[c]{Roll\\-out}} & ConvLSTM-CP-$\Delta Y$ & 22.04$\pm$20.22 & 41.33$\pm$32.04 & 61.00$\pm$42.58 & 77.20$\pm$48.89 \\
    & ConvLSTM-NP-$\Delta Y$ & 21.49$\pm$20.18 & 40.59$\pm$31.83 & 60.15$\pm$42.34 & 76.31$\pm$48.71 \\
    & ConvLSTM-CP-$Y$ & 13.94$\pm$1.59 & 14.89$\pm$1.65 & 16.62$\pm$1.80 & 18.87$\pm$2.09 \\
    & ConvLSTM-NP-$Y$ & 13.30$\pm$1.62 & 14.21$\pm$1.82 & 15.73$\pm$2.27 & 17.34$\pm$2.72 \\
    & \textbf{Alt-ConvLSTM} & \textbf{3.38$\pm$2.08} & \textbf{4.94$\pm$2.98} & \textbf{6.52$\pm$3.70} & \textbf{7.92$\pm$4.11} \\
    \bottomrule
  \end{tabular}
\end{table}

\begin{figure}[t]
    \centering
    \begin{subfigure}[t]{0.16\textwidth}
        \centering
        \includegraphics[width=0.94\linewidth, trim=150 50 150 50, clip]{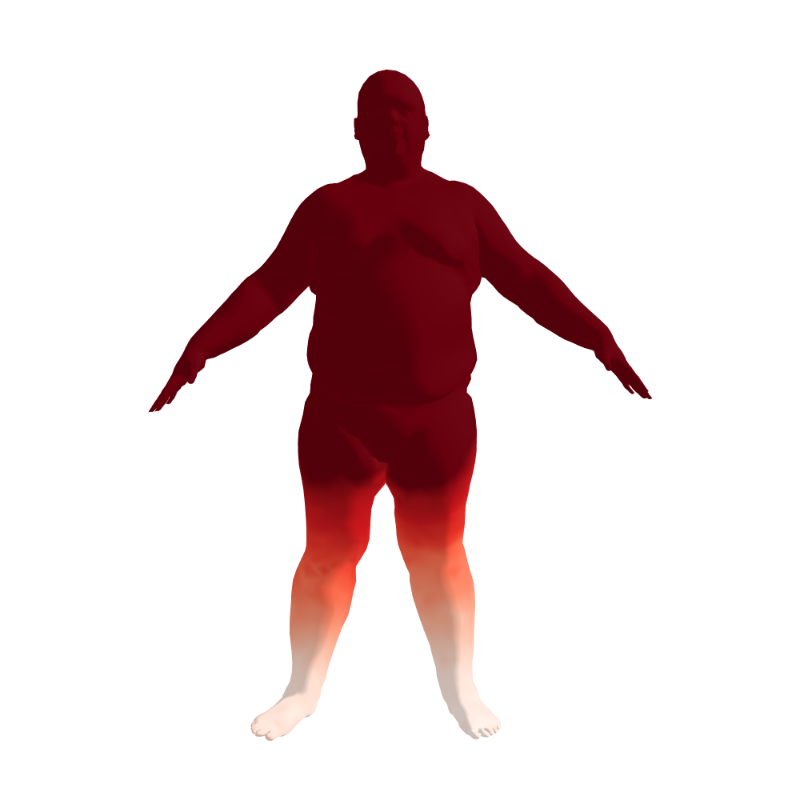}
        \scriptsize{ConvLSTM-CP-$\Delta Y$}
    \end{subfigure}
    \begin{subfigure}[t]{0.16\textwidth}
        \centering
        \includegraphics[width=0.94\linewidth, trim=150 50 150 50, clip]{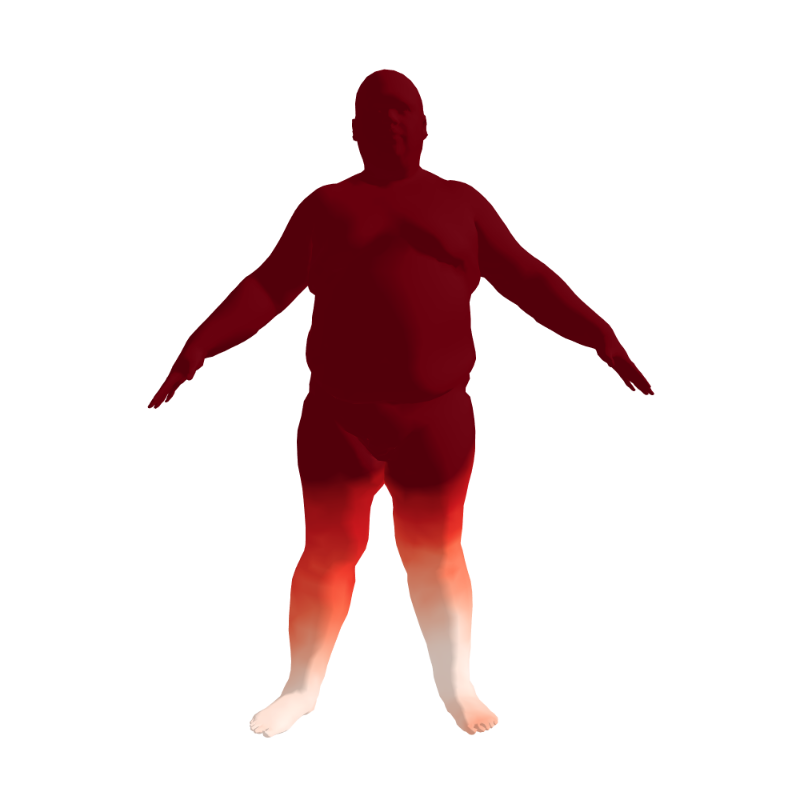}
        \scriptsize{ConvLSTM-NP-$\Delta Y$}
    \end{subfigure}
    \begin{subfigure}[t]{0.16\textwidth}
        \centering
        \includegraphics[width=0.94\linewidth, trim=150 50 150 50, clip]{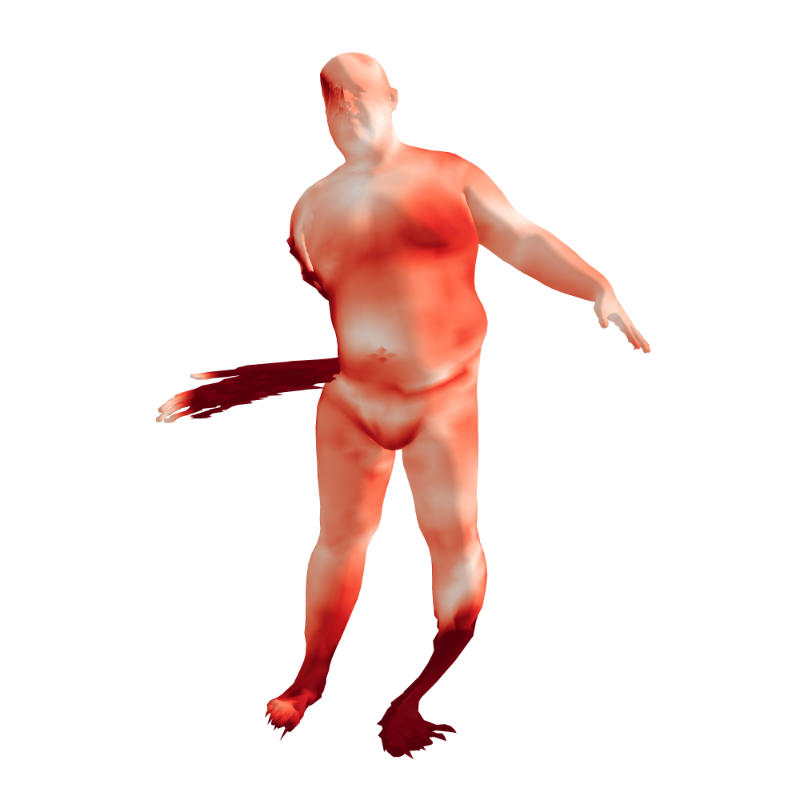}
        \scriptsize{ConvLSTM-CP-$Y$}
    \end{subfigure}
    \begin{subfigure}[t]{0.16\textwidth}
        \centering
        \includegraphics[width=0.94\linewidth, trim=150 50 150 50, clip]{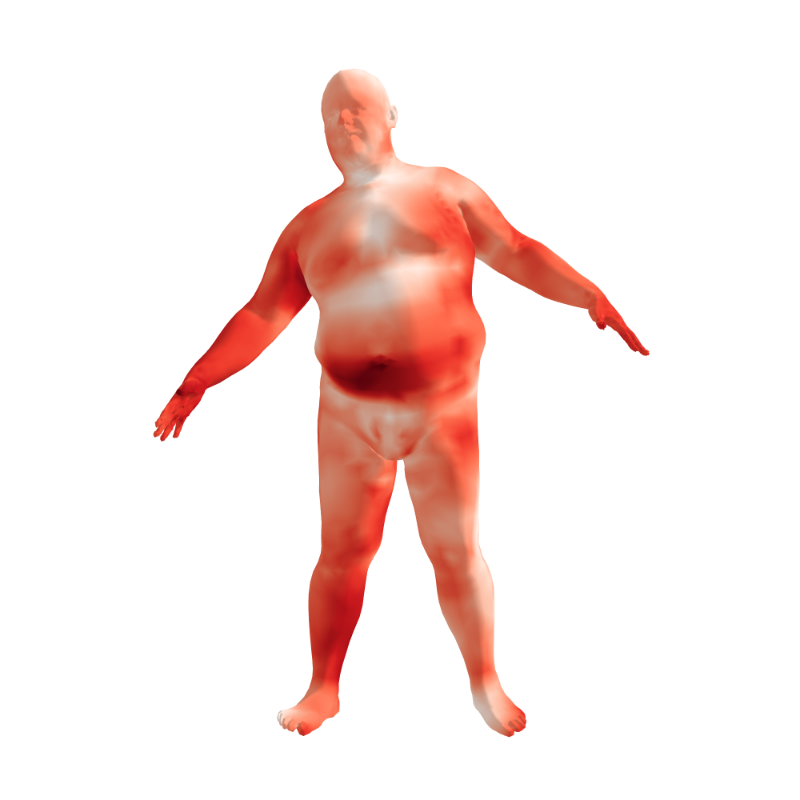}
        \scriptsize{ConvLSTM-NP-$Y$}
    \end{subfigure}
     \begin{subfigure}[t]{0.16\textwidth}
        \centering
        \includegraphics[width=0.94\linewidth, trim=150 50 150 50, clip]{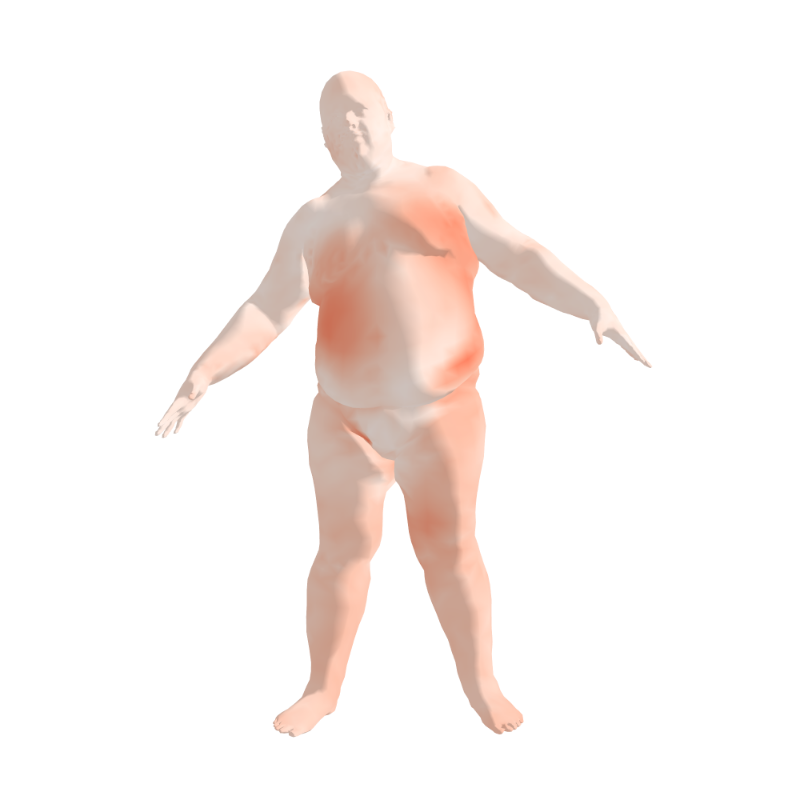}
        \scriptsize{Alt-ConvLSTM}
    \end{subfigure}
     \begin{subfigure}[t]{0.16\textwidth}
        \centering
        \includegraphics[width=0.94\linewidth, trim=150 50 150 50, clip]{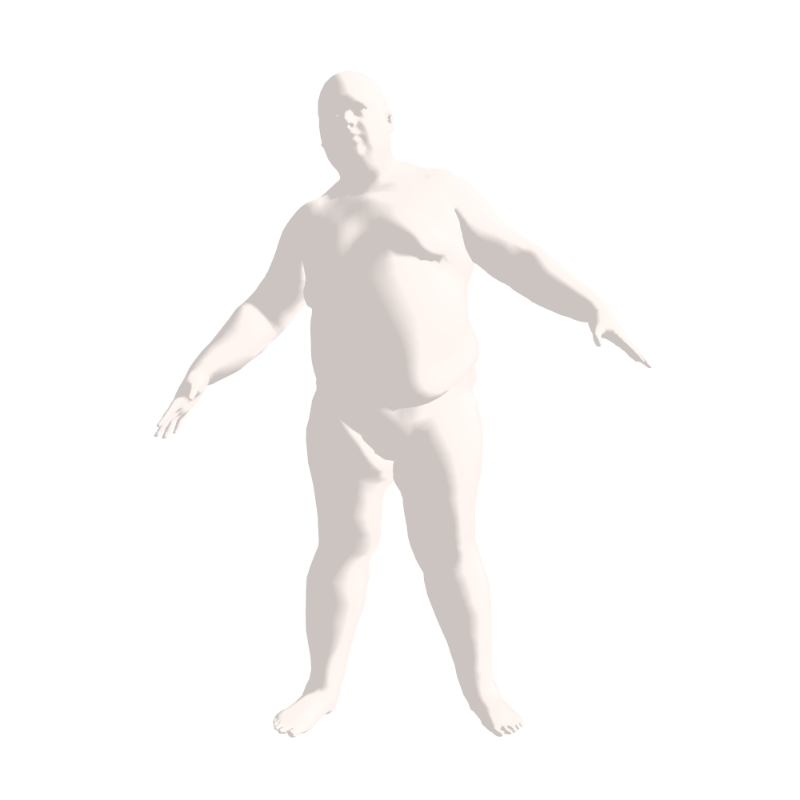}
        \scriptsize{Ground truth}
    \end{subfigure}
    \caption{50-step roll-out simulation results on a known motion with per-vertex errors in red. {\color{Sepia}Cardinal red} indicates an error $\geqslant$5cm.
    Please refer to the supplementary material and video for more results.
    }
    \label{fig:exp_results_known_motions}
\end{figure}

\textbf{Network.}
We build an Alt-ConvLSTM network as described in Sec.~\ref{sec:model} with 5 stacked hidden layers of 32, 64, 128, 64, 32 channels, an output layer of 3 channels, and a skip connection from the input to the output layer by letting
\begin{equation}
    \gY_t^{out}
    = \gY_{t-1}^{out} + \gC_t^{out} + \Delta\rmX_t
    = \gY_{t-1}^{out} + \gC_t^{out} + \rmX_t-\rmX_{t-1}.
\end{equation}

At the beginning when $t=0$, we initialize the output layer with cell accumulation term $\gY_0^{out} = \rmY_0$ and cell output $\gC_0^{out} = 0$, which are the initial vertex coordinates and zero velocities respectively. All other cell states in the network are initialized with zero values.
At each timestep $t\in\{1,\cdots,T\}$, the network takes as input a stacked tensor $[\rmX_t, \rmX_t-\rmX_{t-1}, \rmY_0-\rmX_0]$ representing the external perturbation caused by human subjective intentions.
In the stacked input tensor, $\rmX_t$ is the current body pose, $\rmX_t-\rmX_{t-1}$ is the change in the pose that applies a force on the soft tissue, and $\rmY_0-\rmX_0$ indicates where the force originates beneath the tissue surface.
The network outputs a tensor of vertex velocities $\gC_t^{out} = \widehat{\Delta \rmY}_t$, so the final predicted positions of all vertices are  $\hat{\rmY}_t = \rmY_0 + \sum_{\tau=1}^t \widehat{\Delta \rmY}_\tau$.

\textbf{Loss function.}
We set the loss function to be the mean vertex error in $L_2$-norm across the sequence
\begin{equation}\label{eq:loss}
    \gL(\rmY,\hat{\rmY}) = \frac{1}{T|V|}\sum_{t=1}^T \sum_{i=1}^{|V|} \|\rmY_t(v_i) - \hat{\rmY}_t(v_i)\|_2,
\end{equation}
where $\rmY_t(v_i), \hat{\rmY}_t(v_i)$ are the scanned and predicted coordinates of vertex $v_i$ at timestep $t$ respectively.

\textbf{Training.}
In the training stage, we use the first 21 timesteps of each training sequence; the first timestep is for $\gY_0^{out}$ initialization and the subsequent 20 timesteps are for network training.
Within training, we input ground truth positions to the last network layer, replacing $\hat{\rmY}_{t-1}$ with $\rmY_{t-1}$ for $\gY_{t-1}^{out}$.
We use an Adam optimizer with an initial learning rate of 0.1 and weight decay of 0.995, and after 1000 epochs both the training loss and the validation loss have gone stable.

\subsection{Evaluation}

For all evaluations we adopt both single-step and roll-out errors as proposed in \cite{RempeDynamics2020}. Both metrics measure the mean error of vertex positions as in the Eq. \ref{eq:loss}; the difference is that single-step error uses ground truth historical positions as in the training stage, but roll-out error uses the predicted ones.
The former measures the prediction capability of a temporal model in ideal cases, while the latter also takes into account error accumulation over time.
All the experiments are conducted on a computer with an Intel i9-9900KF 3.60GHz CPU, 32G RAM, and an Nvidia Titan RTX GPU.

\textbf{Future prediction.}
As we train the network with the initial state and the subsequent 20 timesteps, we test its prediction accuracy on the subsequent 21-50 timesteps in the training sequences (known motions).
%
We compare our Alt-ConvLSTM with multiple vanilla ConvLSTM variants of similar architectures and comparable parameter amounts.
Specifically, we test vanilla ConvLSTM with convolutional peephole connections (CP) or no peephole connections (NP), and with network output $\gC^{out}$ to be velocities $\Delta \rmY$ (as in \cite{RempeDynamics2020}) or vertex positions $\rmY$ (as in most other works).
We do not test ConvLSTMs with Hadamard peephole connections of $O(n)$ parameters since they cannot fit into our GPU memory.
All networks comprise 5 hidden layers of 32, 64, 128, 64, 32 channels, an output layer of 3 channels, and a skip connection between the input layer and the output layer.
We train them using an Adam optimizer with initial learning rates $10^{-1}, 10^{-2}, 10^{-3}$ and weight decay 0.995 for 1000 epochs and pick the best results.
%
The single-step and roll-out errors are shown in Table \ref{tab:future_prediction}.
The 20-step error on the left is the training loss, where smaller loss indicates better network representability, and the rest columns are future prediction errors.
We also visualize the 50-step simulation results under the roll-out setting in Figure \ref{fig:exp_results_known_motions}.
The per-vertex errors are shown in red, where darker colors indicates larger errors, and the cardinal red means an error $\geqslant$5cm.

All networks above generate simulation results at around 22 fps on this 6890-particle system.
We see that ConvLSTM's with $\Delta \rmY$ outputs fail to learn the motions but only captures some noises, since it cannot extract historical information efficiently from the raw input of past velocities on this system of thousands of particles.
%
%
ConvLSTM's with $\rmY$ outputs has much less training losses and errors, and results generated by ConvLSTM-NP-$Y$ preserves more semantic meanings than ConvLSTM-CP-$Y$.
However, their errors still accumulate rather fast, as their limited network complexity can hardly model all possible state transitions on this complex physical system.
%
%
In contrast, both the training loss and future prediction errors of our Alt-ConvLSTM are far below all the ConvLSTM variants.
Within same amount of parameters, our Alt-ConvLSTM performs exceedingly excellent representability on modeling physical state transitions than vanilla ConvLSTMs and shows a much slower error accumulation over time for both metrics.

\begin{table}[t]
  \caption{Generalization error on unseen motions}
  \label{tab:generalization}
  \centering
  \begin{tabular}{p{4ex}p{17ex}p{9ex}p{9ex}p{9ex}p{9ex}p{9ex}}
    \toprule
    \multirow{2}{*}{\makecell[c]{Error\\type}} & \multirow{2}{*}{Model} & \multicolumn{5}{c}{Per-vertex error: mean$\pm$sd (mm)} \\
    \cmidrule(r){3-7}
    && 10 steps & 20 steps & 30 steps & 40 steps & 50 steps \\
    \midrule
    & SMPL & 19.40$\pm$3.17 & 19.42$\pm$3.20 & 19.51$\pm$3.13 & 19.64$\pm$2.97 & 19.86$\pm$2.87 \\
    \midrule
    \multirow{3}{*}{\makecell[c]{Single\\-step}} & ConvLSTM-CP-$Y$ & 4.13$\pm$0.69 & 2.66$\pm$0.51 & 2.31$\pm$0.48 & 2.22$\pm$0.46 & 2.69$\pm$0.64 \\
    & ConvLSTM-NP-$Y$ & 4.70$\pm$0.76 & 3.45$\pm$0.62 & 3.12$\pm$0.57 & 3.03$\pm$0.54 & 3.05$\pm$0.54 \\
    & \textbf{Alt-ConvLSTM} & \textbf{0.97$\pm$0.67} & \textbf{1.00$\pm$0.55} & \textbf{1.08$\pm$0.52} & \textbf{1.17$\pm$0.48} & \textbf{1.29$\pm$0.51} \\
    \midrule
    \multirow{3}{*}{\makecell[c]{Roll\\-out}} & ConvLSTM-CP-$Y$ & 14.45$\pm$2.17 & 15.24$\pm$2.32 & 16.10$\pm$2.33 & 17.67$\pm$2.33 & 20.07$\pm$2.81 \\
    & ConvLSTM-NP-$Y$ & 14.06$\pm$2.17 & 14.51$\pm$2.55 & 15.27$\pm$2.90 & 16.60$\pm$3.17 & 18.18$\pm$3.40 \\
    & \textbf{Alt-ConvLSTM} & \textbf{2.32$\pm$1.71} & \textbf{3.70$\pm$2.51} & \textbf{5.18$\pm$3.03} & \textbf{6.64$\pm$3.24} & \textbf{8.14$\pm$3.28} \\
    \bottomrule
  \end{tabular}
\end{table}

\begin{figure}[t]
    \centering
    \begin{subfigure}[t]{\textwidth}
        \centering
        \begin{minipage}{0.16\textwidth}
            \flushleft
            ConvLSTM-ConvP-$Y$
        \end{minipage}
        \begin{minipage}{0.72\textwidth}
        \includegraphics[width=0.19\linewidth, trim=200 50 200 50, clip]{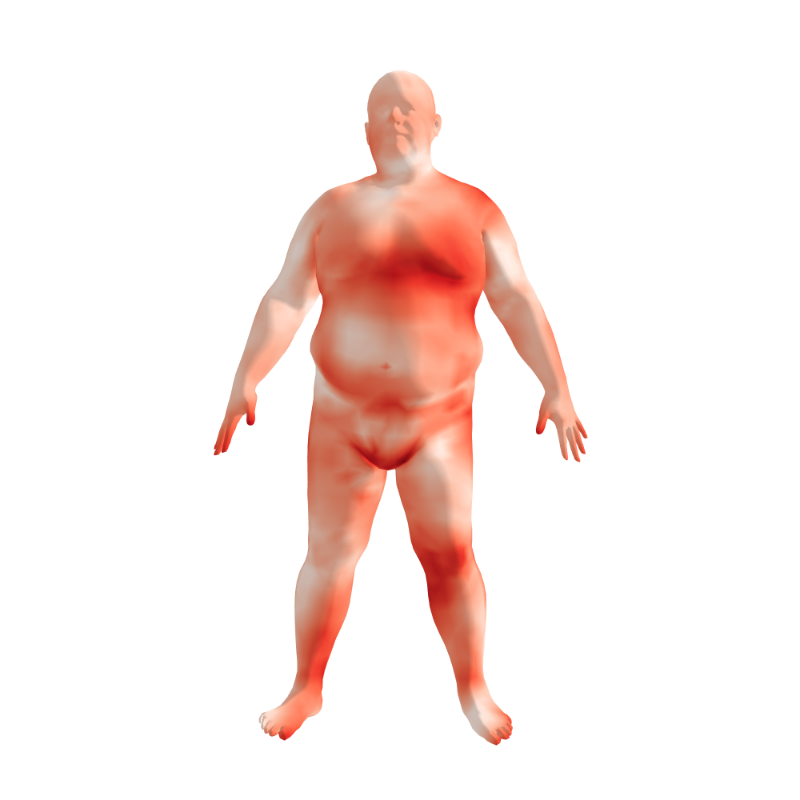}
        \includegraphics[width=0.19\linewidth, trim=200 50 200 50, clip]{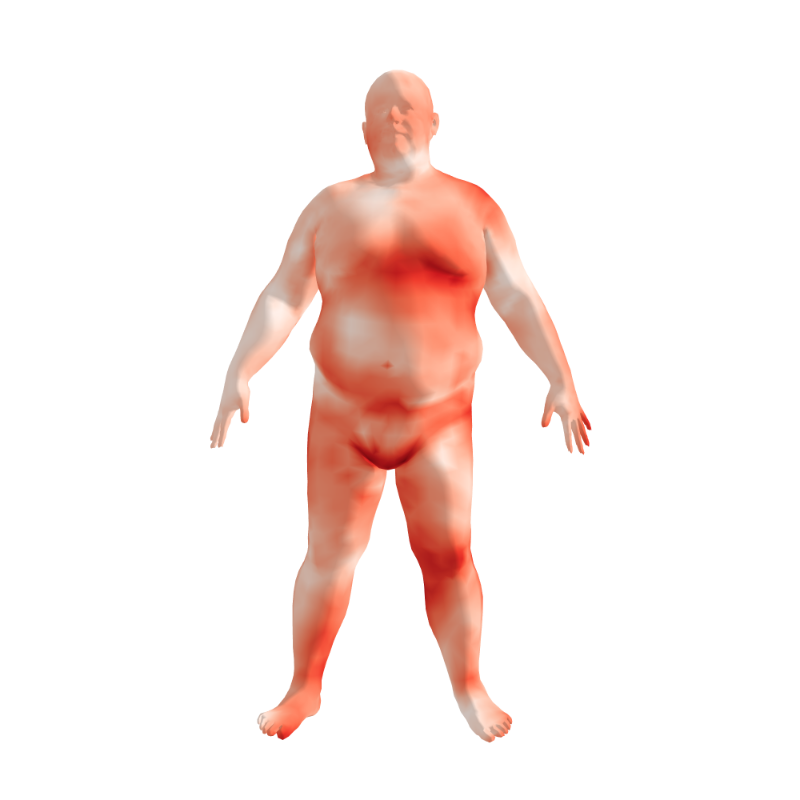}
        \includegraphics[width=0.19\linewidth, trim=200 50 200 50, clip]{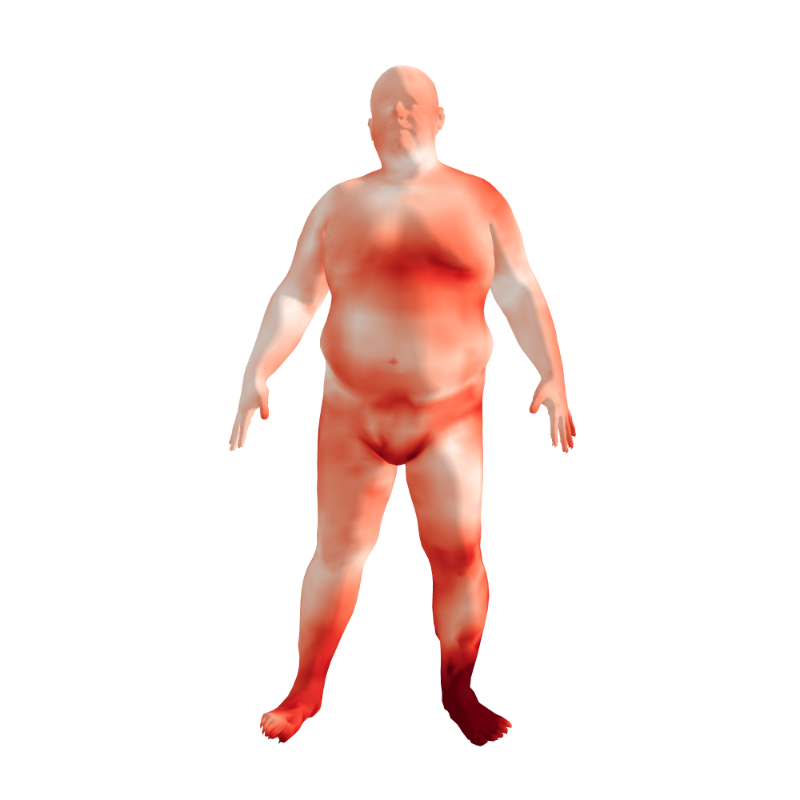}
        \includegraphics[width=0.19\linewidth, trim=200 50 200 50, clip]{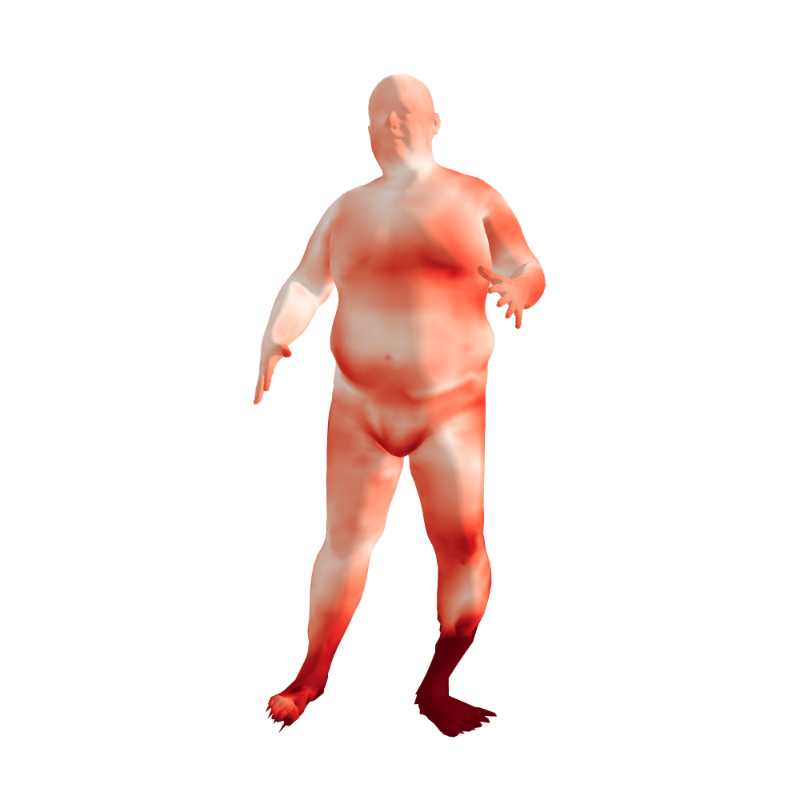}
        \includegraphics[width=0.19\linewidth, trim=200 50 200 50, clip]{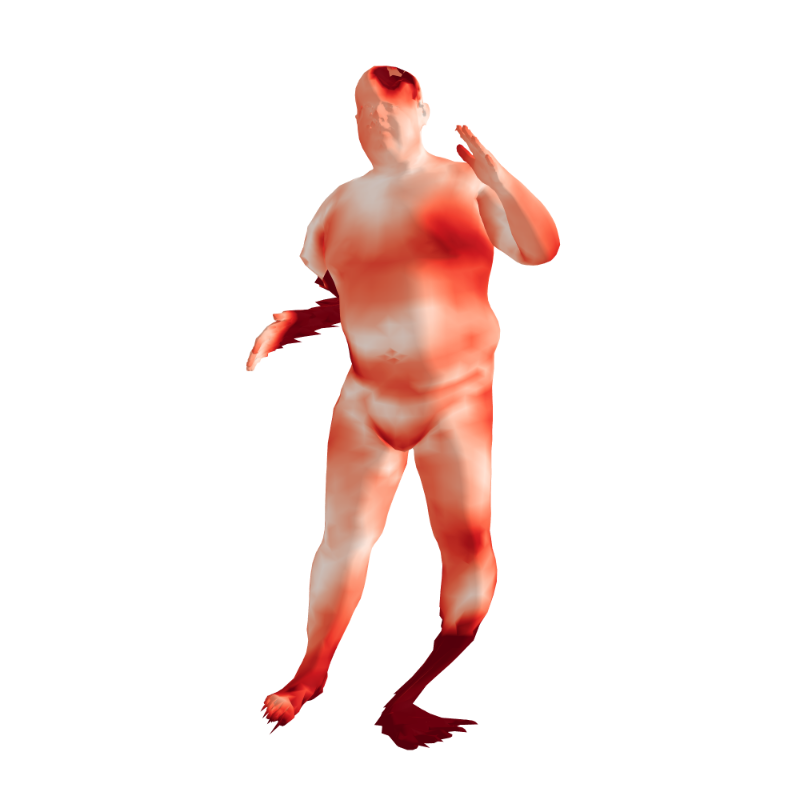}
        \end{minipage}
    \end{subfigure}
    \begin{subfigure}[t]{\textwidth}
        \centering
        \begin{minipage}{0.16\textwidth}
            \flushleft
            ConvLSTM-NP-$Y$
        \end{minipage}
        \begin{minipage}{0.72\textwidth}
        \includegraphics[width=0.19\linewidth, trim=200 50 200 50, clip]{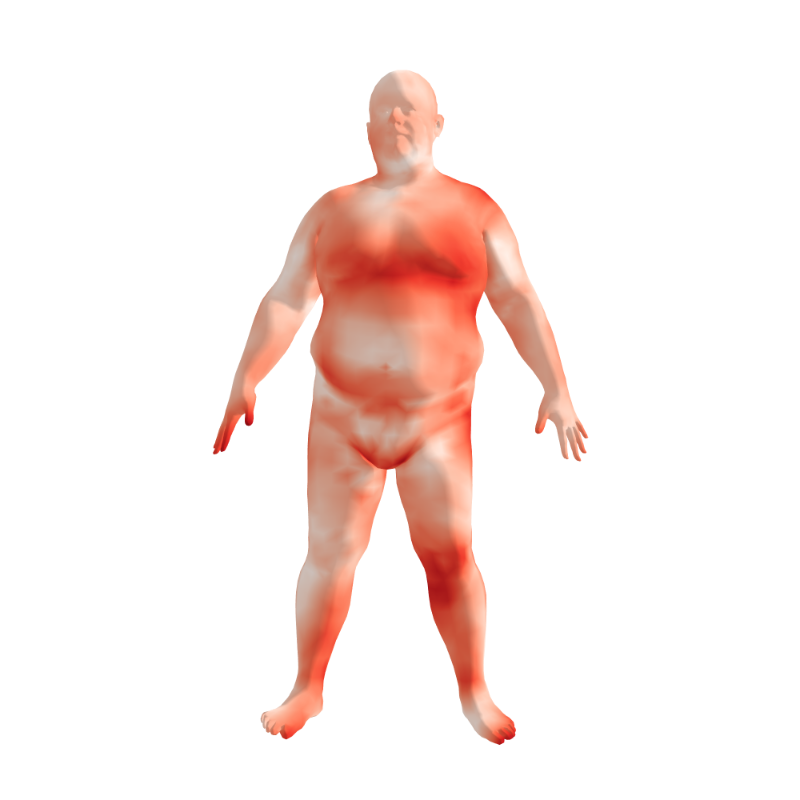}
        \includegraphics[width=0.19\linewidth, trim=200 50 200 50, clip]{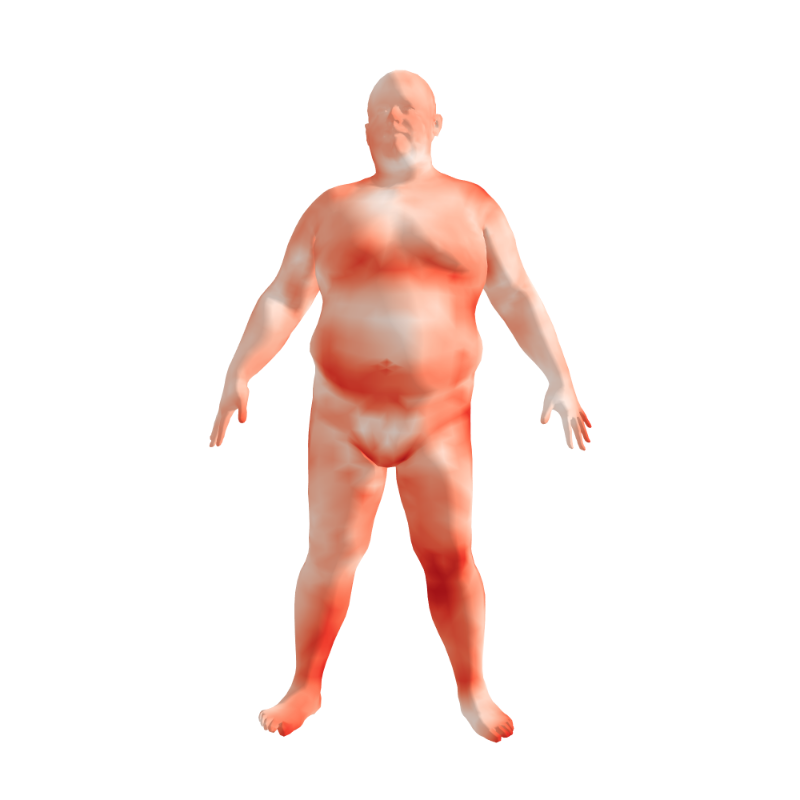}
        \includegraphics[width=0.19\linewidth, trim=200 50 200 50, clip]{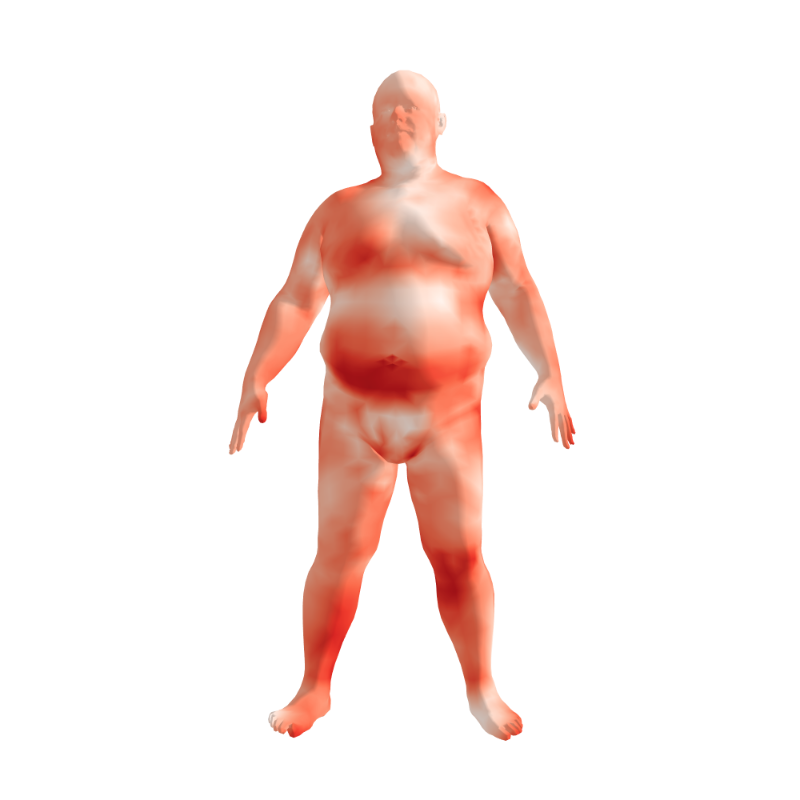}
        \includegraphics[width=0.19\linewidth, trim=200 50 200 50, clip]{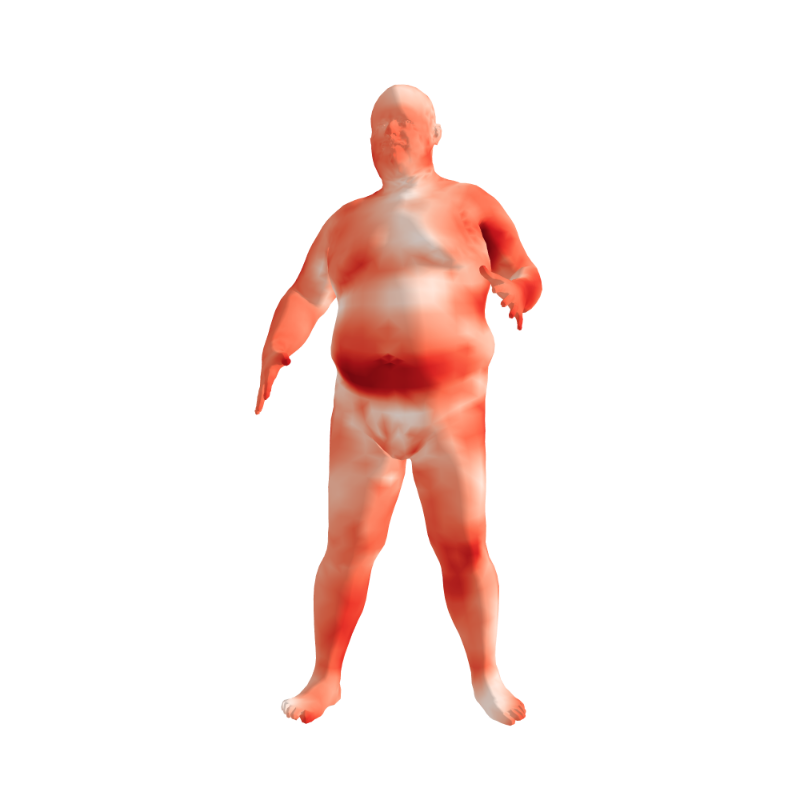}
        \includegraphics[width=0.19\linewidth, trim=200 50 200 50, clip]{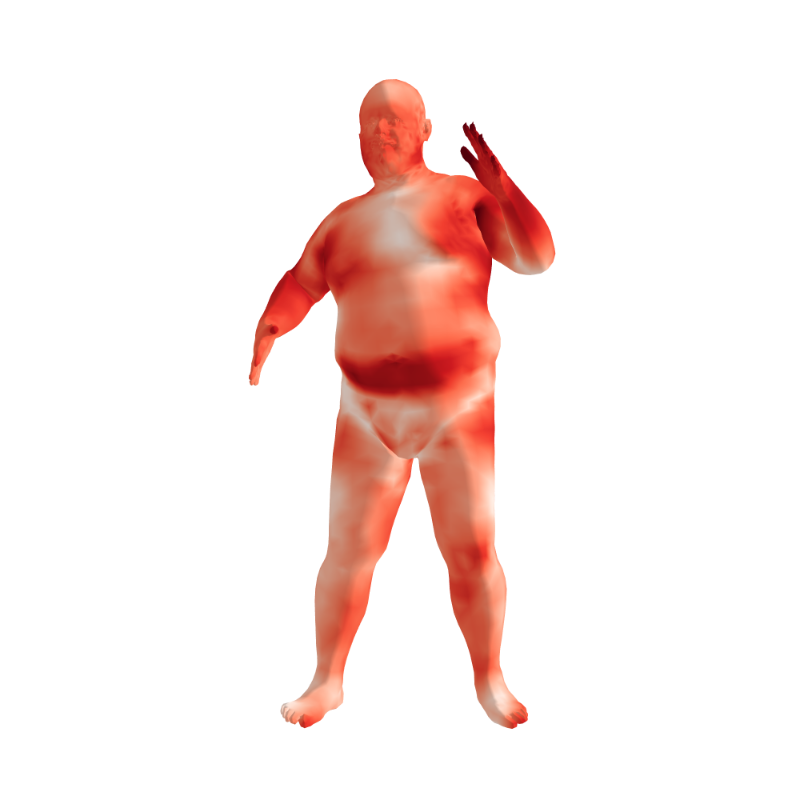}
        \end{minipage}
    \end{subfigure}
    \begin{subfigure}[t]{\textwidth}
        \centering
        \begin{minipage}{0.16\textwidth}
            \flushleft
            Alt-ConvLSTM
        \end{minipage}
        \begin{minipage}{0.72\textwidth}
        \includegraphics[width=0.19\linewidth, trim=200 50 200 50, clip]{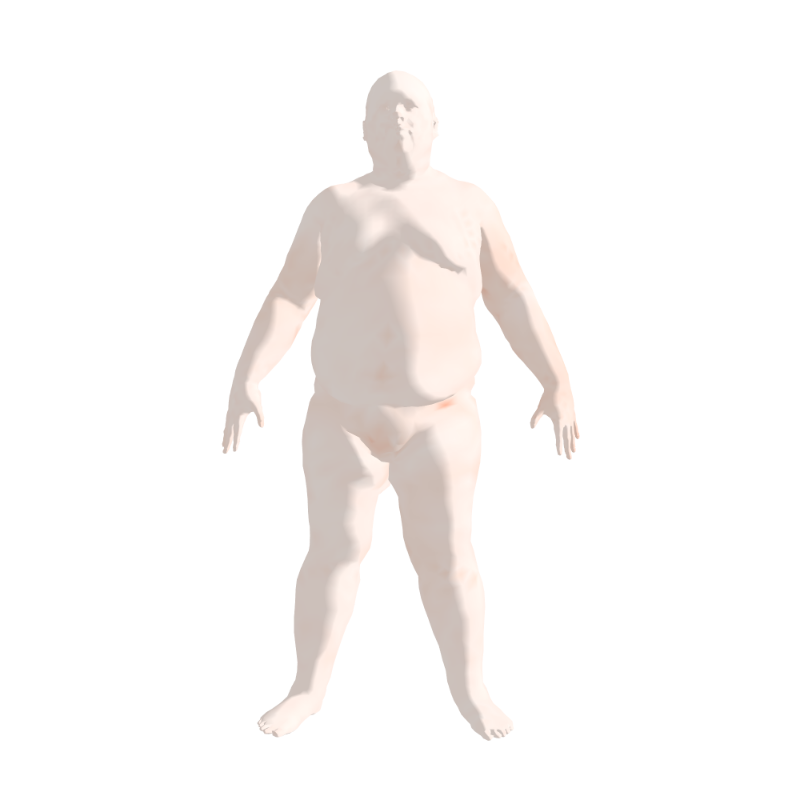}
        \includegraphics[width=0.19\linewidth, trim=200 50 200 50, clip]{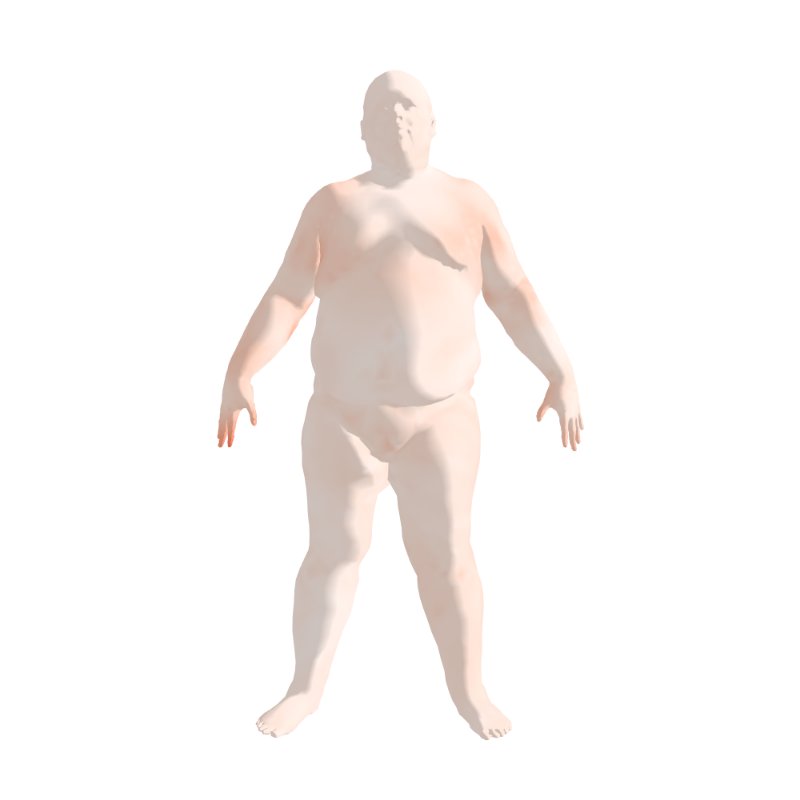}
        \includegraphics[width=0.19\linewidth, trim=200 50 200 50, clip]{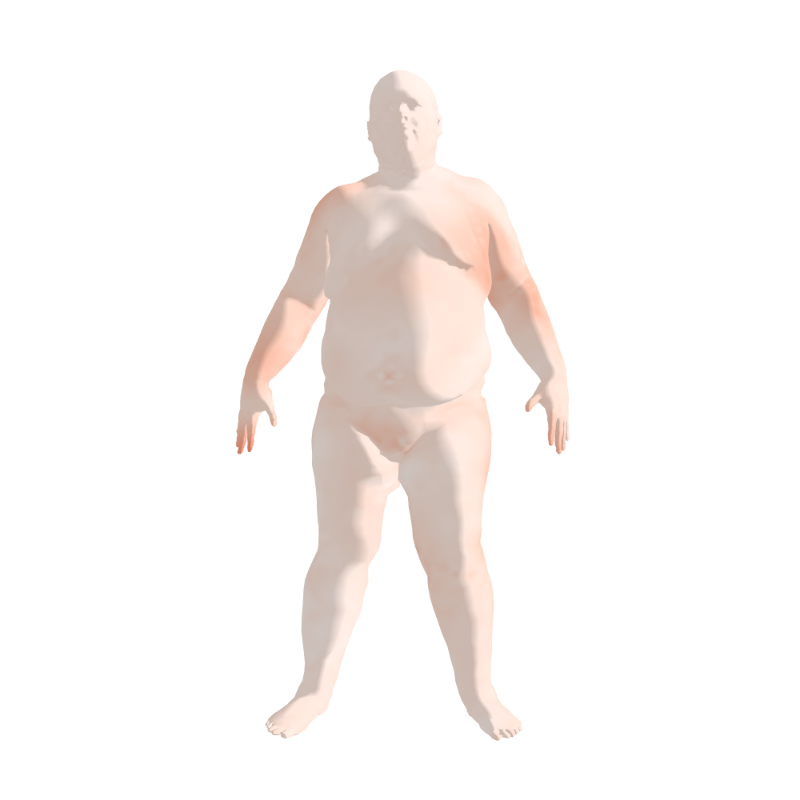}
        \includegraphics[width=0.19\linewidth, trim=200 50 200 50, clip]{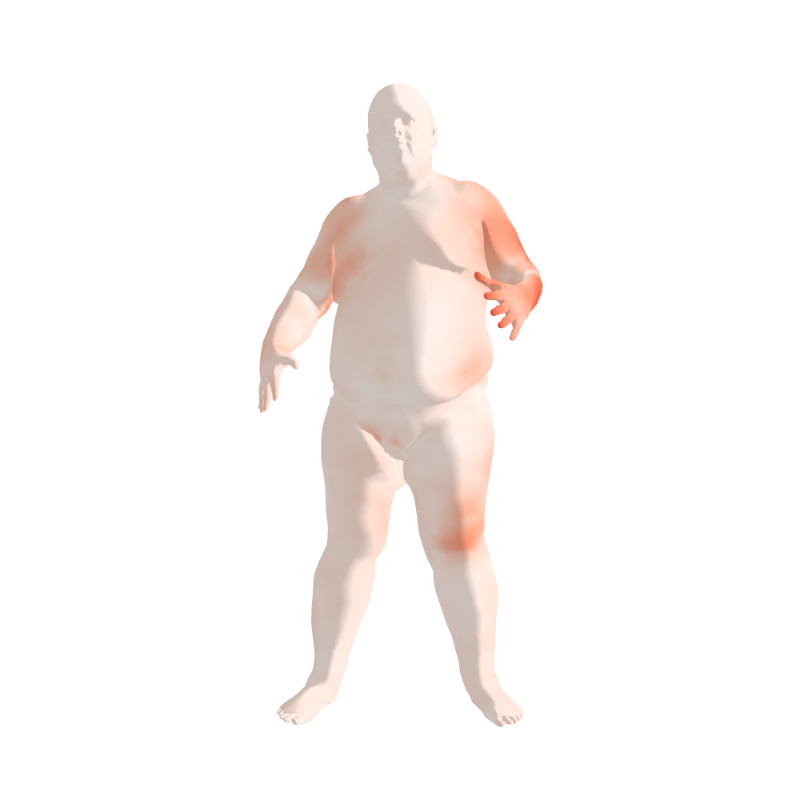}
        \includegraphics[width=0.19\linewidth, trim=200 50 200 50, clip]{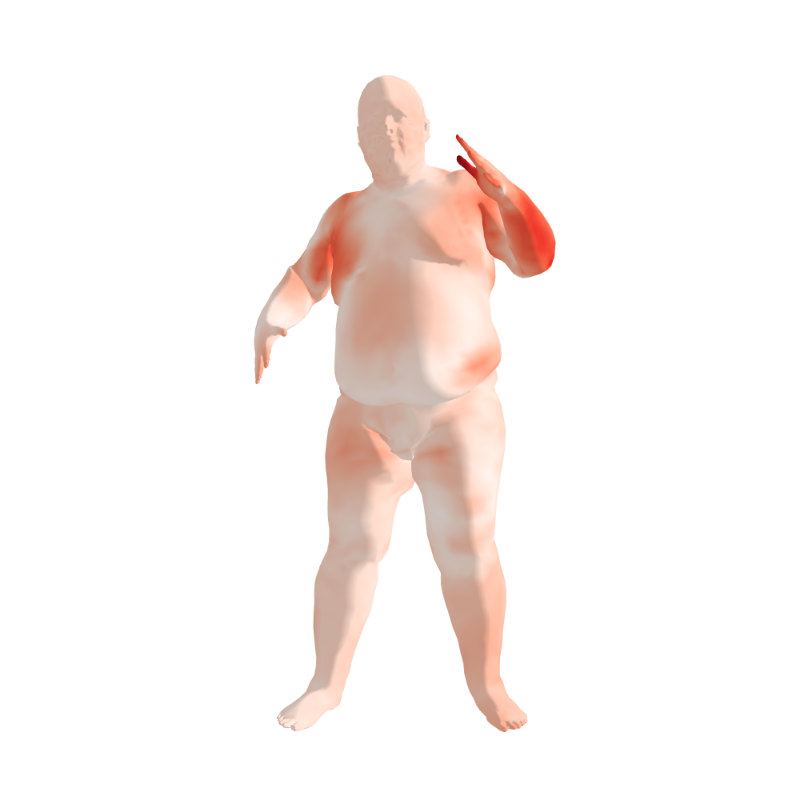}
        \end{minipage}
    \end{subfigure}
    \begin{subfigure}[t]{\textwidth}
        \centering
        \begin{minipage}{0.16\textwidth}
            Ground truth
        \end{minipage}
        \begin{minipage}{0.72\textwidth}
        \includegraphics[width=0.19\linewidth, trim=200 50 200 50, clip]{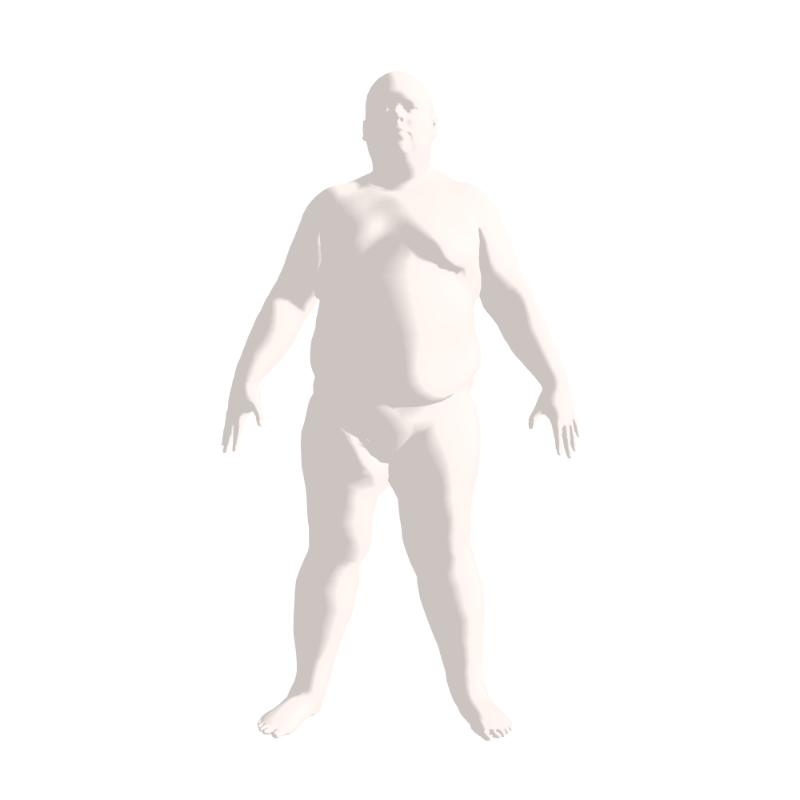}
        \includegraphics[width=0.19\linewidth, trim=200 50 200 50, clip]{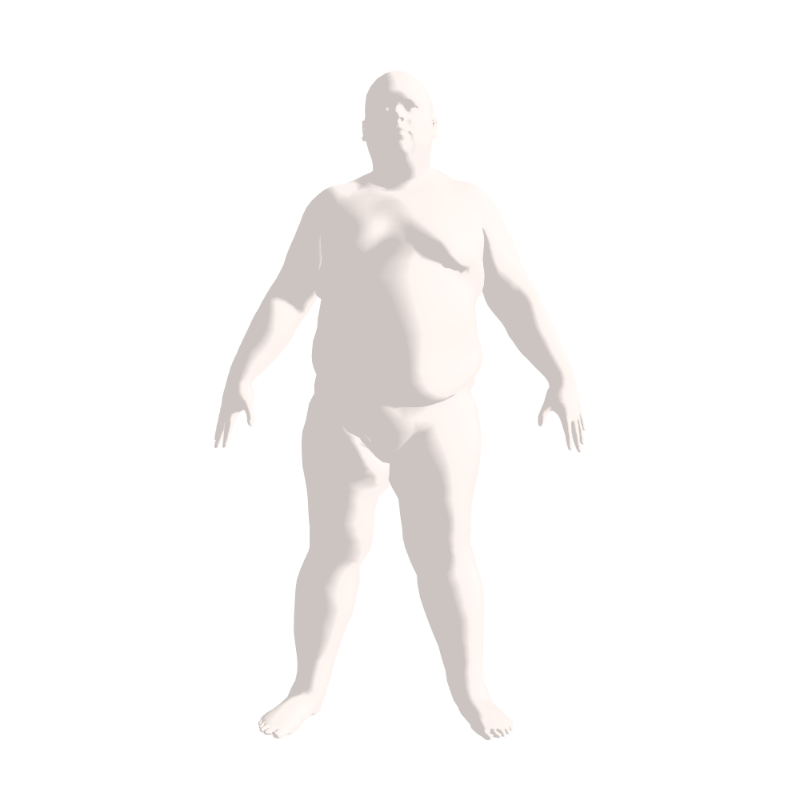}
        \includegraphics[width=0.19\linewidth, trim=200 50 200 50, clip]{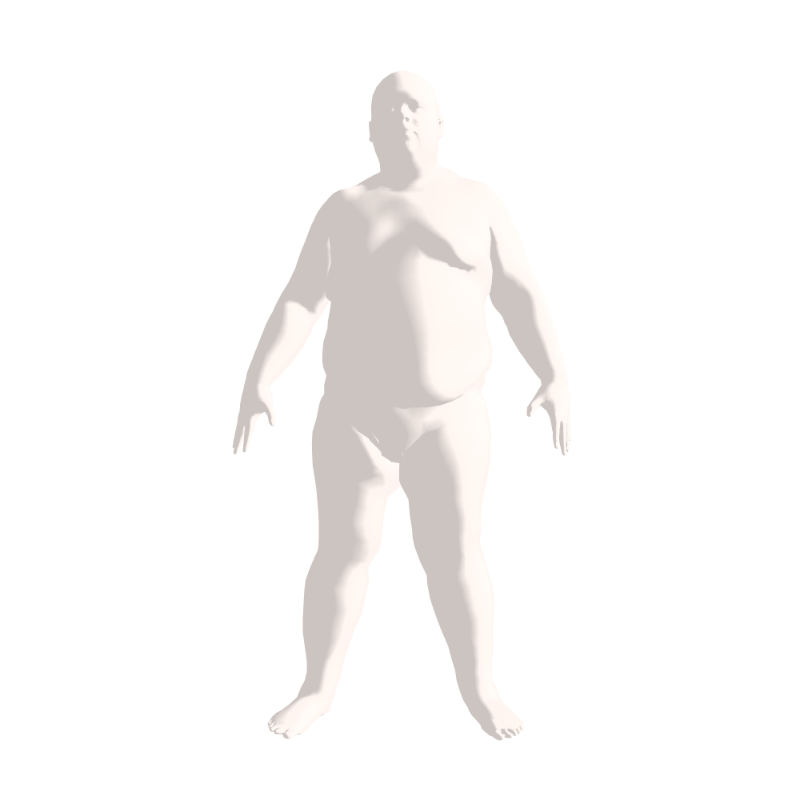}
        \includegraphics[width=0.19\linewidth, trim=200 50 200 50, clip]{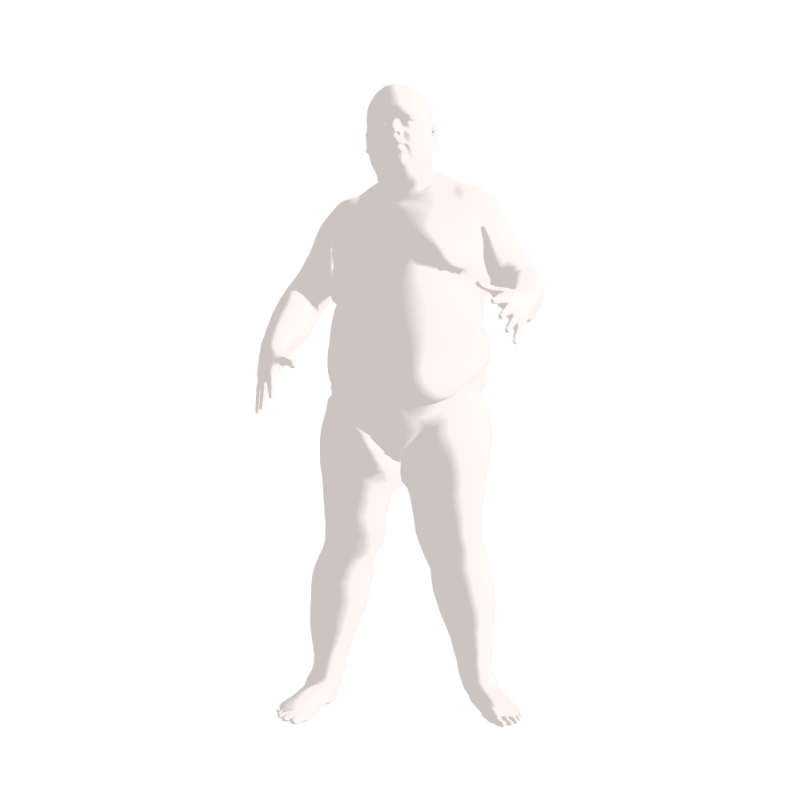}
        \includegraphics[width=0.19\linewidth, trim=200 50 200 50, clip]{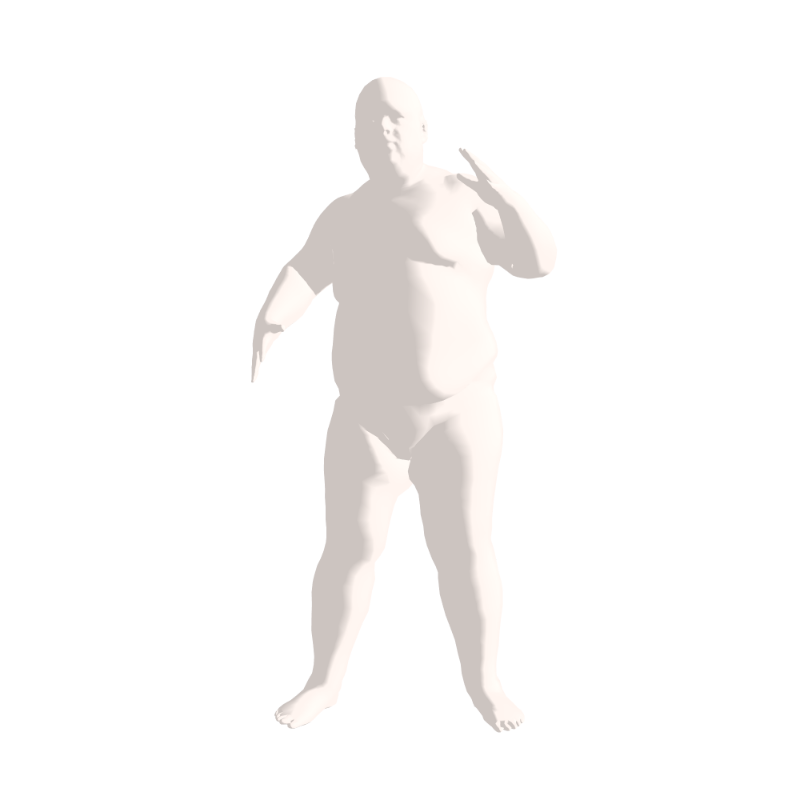}
        \end{minipage}
    \end{subfigure}
    \caption{Simulation results at 10, 20, 30, 40, 50 steps (from left to right) on an unseen motion with per-vertex errors in red. {\color{Sepia}Cardinal red} indicates an error $\geqslant$5cm.
    Please refer to the supplementary material and video for more results.
    }
    \label{fig:exp_results_unseen_motions}
\end{figure}

\textbf{Generalization to new sequences.}
%
We also evaluate our method and ConvLSTMs with $\rmY$ cell outputs on the unseen motions in the test set.
%
The generalization errors are in Table \ref{tab:generalization}, and their simulation results are shown in Figure \ref{fig:exp_results_unseen_motions}.
We see that both single-step and roll-out errors of our Alt-ConvLSTM remains comparably low on these unseen sequences, indicating that our network is indeed modeling some innate material features that are independent of any specific physical states.

As a reference, we also give the average error $6.27\pm1.87$ of an autoencoder-based method \cite{casas2018learning} tested on 3 sequences.
%
%
The single-step errors and the roll-out errors of 10-20 steps show that, for short-period simulations, our Alt-ConvLSTM greatly outperforms\cite{casas2018learning}, while for longer periods, its roll-out error gradually approaches and exceeds \cite{casas2018learning} due to error accumulation.
To stress our superiority, we emphasize that even the state-of-the-art autoencoder has a reconstruction error of at least 2.09 mm on this dataset \cite{santesteban2020softsmpl}, not to mention their additional errors in temporal regressions.

\section{Conclusion and Future Work}

In this paper, we propose the Alt-ConvLSTM network for simulating deformable objects under external forces.
The network structure holds strong physical interpretations of imitating forward Euler updates and modeling force propagation.
The preserved ConvLSTM features also enables the network to handle spatial-temporal inputs and outputs skillfully.
Because of the well-designed structures, the network acts on local regions and has only a small number of parameters, but shows strong capabilities on representing physical state transition.
%
%
%
Our current method still has limitations.
One major limitation is the error accumulation over time, which is inevitable for nearly all temporal prediction methods.
%
Another is the network initialization, as we should further study how to initialize the hidden layers according to the initial positions and velocities.
We will also explore more complicated physical simulation tasks, including learning physical details of higher frequency such as bumps and pits on the mesh faces, or multi-object interactions with collision detection or dynamical interaction graphs.
%
%

\section*{Broader Impact}

Major beneficiaries of this research would be the entertainment industries, including the movie industry and the game industry.
We do not think any individual or organization will be put at disadvantage from this research.
For simple simulations and animations, failure of the system will not lead to any harmful consequences, at least in the visible future.
But if it is further used for mechanical engineering or robotics, we should be cautious of the result of failure of the system which could cause erroneous plannings and controls.
The method is not leveraging biases in the data.



\begin{ack}
This work is supported by the National Natural Science Foundation of China under Grant Nos. 61902210 and 61521002, a research grant from the Beijing Higher Institution Engineering Research Center, and the Tsinghua-Tencent Joint Laboratory for Internet Innovation Technology.
%

\end{ack}

\small
\bibliographystyle{unsrt}
\bibliography{references}

\end{document}